\documentclass[pdflatex,iicol,sn-mathphys-num]{sn-jnl}

\usepackage{graphicx}%
\usepackage{multirow}%
\usepackage{amsmath,amssymb,amsfonts}%
\usepackage{amsthm}%
\usepackage{mathrsfs}%
\usepackage[title]{appendix}%
\usepackage{xcolor}%
\usepackage[normalem]{ulem}
\usepackage{textcomp}%
\usepackage{manyfoot}%
\usepackage{booktabs}%
\usepackage{algorithm}%
\usepackage{algorithmicx}%
\usepackage{algpseudocode}%
\usepackage{listings}%
\usepackage{siunitx}
\usepackage{array}
\usepackage{lipsum}
\usepackage{booktabs}
\usepackage{tabularx}
\usepackage{amsmath}
\usepackage{graphicx}
\usepackage{amssymb}
\usepackage{stfloats}
\usepackage{listings}
\usepackage{array} 
\usepackage{booktabs}
\usepackage{subcaption}
\usepackage{float}
\usepackage{afterpage}
\usepackage{needspace}

\let\orcidlogo\relax
\usepackage{orcidlink}

\theoremstyle{thmstyleone}%
\theoremstyle{thmstyletwo}%

\theoremstyle{thmstylethree}%

\begin{document}

\title[Article Title]{DRA-HTR: Disentangling Representation and Adaptation for Handwritten Text Recognition}


\author[1]{\fnm{Gayan H.} \sur{Pathirage}\orcidlink{0009-0000-9164-487X}}\email{gayan.pathirage@ltu.se}
\equalcont{These authors contributed equally to this work.}

\author[1]{\fnm{Florent} \sur{Imbert}\orcidlink{0009-0008-2136-0829}}\email{florent.imbert@associated.ltu.se}
\equalcont{These authors contributed equally to this work.}

\author[2]{\fnm{Denis} \sur{Coquenet}\orcidlink{0000-0001-5203-9423}}\email{denis.coquenet@irisa.fr}

\author[1]{\fnm{Elisa H.} \sur{Barney Smith}\orcidlink{0000-0003-2039-3844}}\email{elisa.barney@ltu.se}

\affil[1]{\orgdiv{Machine Learning Group}, \orgname{Luleå University of Technology}, \orgaddress{\city{Luleå}, \country{Sweden}}}

\affil[2]{\orgdiv{Univ Rennes}, \orgname{CNRS, IRISA - UMR 6074}, \orgaddress{\city{F-35000 Rennes}, \country{France}}}


\abstract{Handwritten text recognition (HTR), particularly when applied to historical documents, remains challenging due to the scarcity of annotated data and the substantial domain shifts caused by variations in handwriting styles, document degradation, acquisition conditions, and linguistic characteristics. While large amounts of unlabeled target-domain data are often available, effectively exploiting them requires adaptation strategies tailored to the specific requirements of HTR.
We propose DRA-HTR, an unsupervised generative domain adaptation pipeline specifically designed for HTR and built upon the DRANet framework. The proposed approach combines source-domain pretraining, unsupervised adaptation to the visual characteristics of the target domain, relying on the generation of target-style source samples for task-specific training. We further investigate the integration of post-HTR correction to complement visual adaptation with language adaptation, relying on large language models, which we introduce as DRAL-HTR.
The proposed framework is evaluated across seven HTR datasets spanning four languages, using a single set of hyperparameters for all source–target combinations. The recognition model achieves state-of-the-art performance on four datasets and competitive results on the remaining three. In addition, the proposed DRA-HTR establishes a new state-of-the-art for domain adaptation across all seven datasets and substantially improves overall robustness and training stability compared to the original DRANet framework~\cite{Imbert2026}: it reduces the Character Error Rate (CER) in average by 13.52 percentage points, and the average standard deviation decreases from 7.23 to 1.11 percentage points. Finally, post-HTR correction further reduces the CER by 2.16 percentage points.
Code is available at \url{https://github.com/FIorentI/DRA-HTR-Disentangling-Representation-and-Adaptation-for-Handwritten-Text-Recognition}.
}


\maketitle

\section{Introduction}\label{sec_intro}
Handwritten text recognition (HTR) has achieved substantial progress with the development of deep learning, enabling increasingly accurate transcription of handwritten documents. However, HTR performance remains highly dependent on the similarity between the characteristics of the training and evaluation data. Variations in handwriting styles, acquisition conditions, document degradation, writing instruments, historical periods, or languages can introduce significant discrepancies between training and test data. Consequently, a recognition model that performs well on one dataset may experience a substantial loss of accuracy when deployed on a new corpus~\cite{domain_shift}.

A straightforward solution consists of collecting and manually transcribing representative samples from each new domain or collection and subsequently fine-tuning or retraining the recognition model. In practice, this strategy is difficult to scale. Producing line-level transcriptions is particularly time-consuming and may additionally require expertise in historical scripts, languages, or document collections. At the same time, large quantities of untranscribed document images are often readily available. Exploiting these unlabeled images therefore represents an important opportunity for adapting recognition systems while limiting the amount of manual annotation required.

Unsupervised domain adaptation (DA) provides a suitable framework for this setting~\cite{Imbert2026}. Given a model trained from labeled data in a source domain, the objective is to exploit unlabeled samples from a target domain to reduce the discrepancy between the two distributions. For HTR, this discrepancy is not limited to handwriting appearance. Recognition depends both on the visual representation of the written content and on the linguistic regularities that constrain possible transcription sequences. Effective adaptation to a new collection should therefore account for changes in visual characteristics while preserving the original textual content.

In this work, we build upon the DA framework proposed by~\cite{Imbert2026}, originally designed for handwritten character classification and also applied to HTR. We refine the quality of the generated images, improving the recognition accuracy, robustness and generalization across domains without requiring annotated target-domain data and further extend this framework with post-HTR correction. 
The proposed approach transforms labeled source-domain samples toward the visual characteristics observed in an unlabeled target domain. Consequently, the resulting target-styled samples retain their source-domain labels and can be used as supervised training data for the HTR model.
This strategy allows target-domain visual information to be incorporated into the recognition process without requiring manual transcription of the target images.

\begin{figure}[t]
\centering
\includegraphics[width=\linewidth]{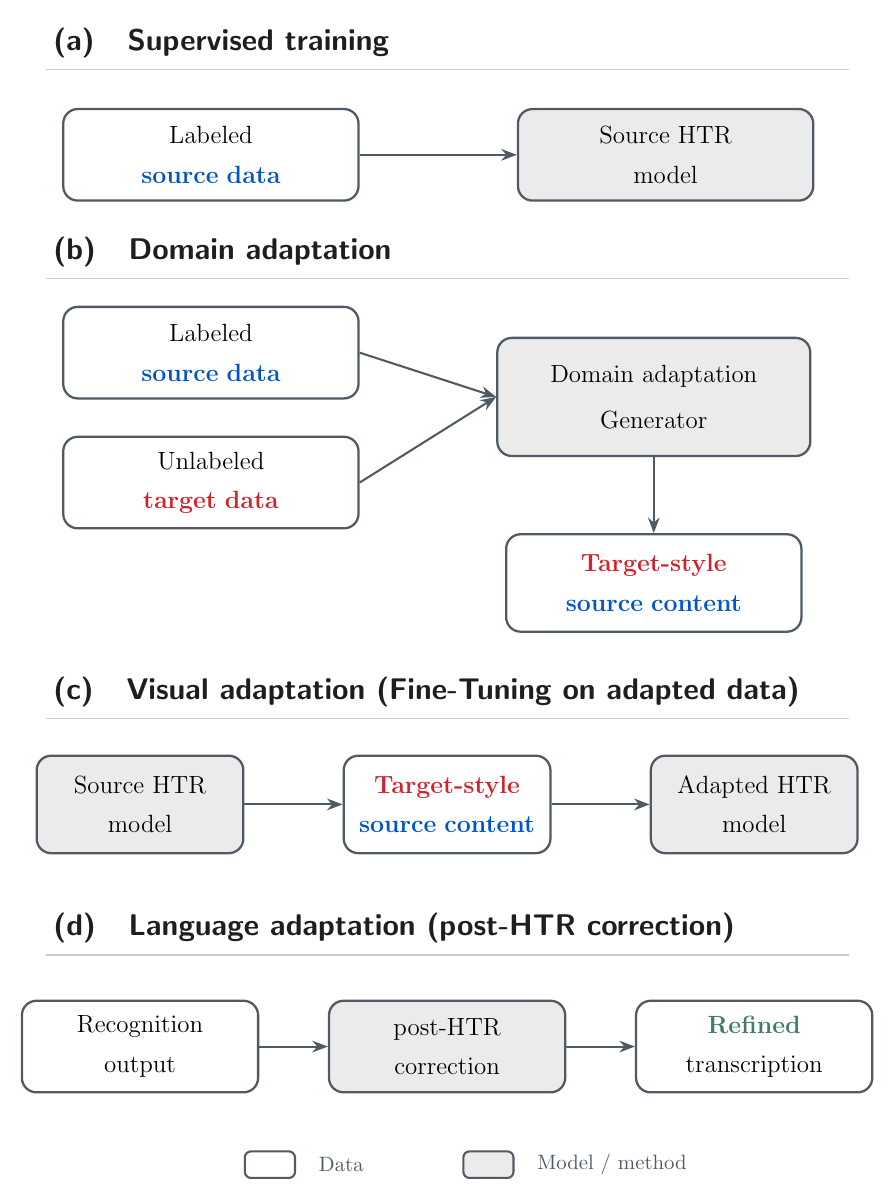}
\caption{Overview of the DA pipeline. Step (a): supervised training of the HTR model on labeled source-domain data. Step (b): unsupervised visual DA using source and unlabeled target data to generate samples in the target style. Step (c): training of the HTR model using the generated target-style images, with their source transcriptions. Step (d): post-HTR correction using Qwen to refine the recognized text by leveraging linguistic context.}
\label{fig_overview_method}
\end{figure}

The overall framework, illustrated in Figure~\ref{fig_overview_method}, is organized into three stages. First, an HTR model is trained on labeled source-domain data to establish a strong recognition baseline. Second, visual DA is performed using labeled source images together with unlabeled target images to learn a generative transformation toward the target appearance. Third, the source images transformed into the target style, together with their original transcriptions, are used to fine-tune the HTR model to the target domain. Post-HTR correction is further incorporated into the recognition pipeline to exploit linguistic context and improve transcription consistency.

To assess the robustness of the proposed framework, we conduct experiments across 7 datasets covering 4 different languages. This evaluation setting introduces substantial variations in handwriting styles, document characteristics, and linguistic content, allowing us to investigate both the visual generalization capabilities of the adaptation strategy and the contribution of post-HTR correction.
The main contributions are as follows:
\begin{itemize}
\item We introduce DRA-HTR, an improved DA framework specifically designed for HTR. Building upon previous work, we integrate a ConvNeXtV2-based encoder to better capture complex handwriting features, redesign the separator and generator, and introduce a progressive training strategy with delayed HTR supervision. These improvements enhance the quality of generated target-style handwritten images while preserving textual content, leading to more effective cross-domain adaptation and higher recognition accuracy.
\item We extend DRA-HTR with LLM-based post-HTR correction, resulting in DRAL-HTR. We propose hypothesis-based and context-based correction strategies that leverage multiple transcription hypotheses and neighboring lines to improve recognition accuracy beyond visual information alone.
\item We conduct extensive experiments on 7 datasets spanning 4 languages, demonstrating the effectiveness and generalization capabilities of our framework. Comprehensive ablation studies further validate the individual contributions.
\end{itemize}

The remainder of this paper is organized as follows. Section~\ref{sec_sota} reviews related work on HTR, DA, and post-HTR correction. Section~\ref{sec_method} presents the proposed framework and its individual components. Section~\ref{sec_expe} describes the experimental setup and reports quantitative and qualitative results. Finally, Section~\ref{sec_conclu} concludes the paper and discusses directions for future research.

\section{Related Work}\label{sec_sota}

\subsection{Handwritten Text Recognition}
HTR aims to transcribe handwritten text line images into machine-readable text. At the line level, HTR systems generally rely on neural architectures that jointly learn visual representations and sequential dependencies, avoiding the explicit segmentation of text into individual characters. Convolutional Recurrent Neural Networks (CRNN) remain widely used for this task~\cite{Retsinas22,Tarride24,Corbille25}. In these architectures, convolutional layers extract visual features from the input line image, while recurrent layers, typically based on Long Short-Term Memory (LSTM) units, model dependencies along the writing direction. The resulting sequence is commonly decoded using Connectionist Temporal Classification (CTC) \cite{Graves06}, which enables end-to-end training without requiring character-level alignment between the image and its transcription. In~\cite{Chinthaginjala2024}, Bidirectional Gated Recurrent Unit (BGRU) layers were employed for sequence modeling, and CTC decoding was combined with a character-level 9-gram language model to refine the recognition predictions.

Transformer-based architectures have also been investigated for HTR~\cite{barrere24,LiL23}. By relying on attention mechanisms, these approaches can model long-range dependencies without recurrent operations and have demonstrated strong recognition performance. MetaWriter~\cite{Gu_2025_CVPR} combines a Fully Convolutional Network (FCN) encoder with a Transformer decoder and uses meta-learned prompt tuning for personalized HTR. Similarly, Meta-DAN~\cite{COQUENET2026113373} employs an FCN encoder and a Transformer decoder for efficient page-level text recognition. However, the effectiveness of transformer-based architectures generally depends on large training sets and substantial computational resources. CRNN-based architectures therefore remain particularly relevant when dealing with historical collections or specialized document domains for which annotated data are limited. For this reason, we adopt a CRNN-based architecture as the HTR model in our proposed DA pipeline.

Despite the progress achieved by these architectures, their performance remains strongly dependent on the correspondence between the training and evaluation distributions. Changes in handwriting style, acquisition conditions, document degradation, background appearance, or historical period can substantially affect the visual representation of handwritten text. Consequently, an HTR model trained on one collection may experience a significant performance degradation when directly applied to another. This limitation motivates the use of DA techniques to exploit unlabeled data from a new document collection.

\subsection{Representation Learning for Image Generation}

Learning effective feature representations is important for controlling the appearance and structural properties of generated images. One approach to achieving such control is feature-wise linear modulation (FiLM), which was introduced for visual question answering, where a question embedding modulates visual feature maps to guide the reasoning process~\cite{Perez2018FiLM}. Beyond visual reasoning, FiLM has been employed to condition image-generation networks. In~\cite{Ren2021QSpace}, continuous \(q\)-space information modulates instance-normalized generator features to synthesize diffusion-weighted brain MR images with specified diffusion characteristics. Similarly, in~\cite{Ren2021Segmentation}, anatomical segmentation embeddings modulate normalized generator features throughout unpaired medical image harmonization, improving image fidelity while preserving anatomical structures. These studies demonstrate that FiLM provides fine-grained control over intermediate features and supports the learning of condition-aware representations, thereby improving the fidelity and structural consistency of generated images.

\subsection{Domain Adaptation}\label{sec_sota_DA}
DA addresses distribution shifts between a labeled source domain and a target domain whose data follow a different distribution. Rather than requiring the complete annotation of the target collection, DA seeks to transfer the knowledge acquired from the source domain while accounting for the characteristics of the target data.

Different settings can be considered depending on the amount of supervision available in the target domain. Supervised and semi-supervised DA respectively assume fully or partially annotated target samples, whereas Unsupervised DA relies exclusively on labeled source data and unlabeled target data. The latter setting is particularly attractive for document analysis, where producing line-level transcriptions for every new collection is costly and time-consuming.

Existing DA strategies can broadly operate either on learned representations or directly on the appearance of the input data. Feature-level approaches seek to learn representations whose distributions are similar across domains. This can be achieved by explicitly minimizing statistical discrepancies between source and target features~\cite{shen2018wassersteindistanceguidedrepresentation}, or through adversarial objectives in which a domain discriminator encourages the feature extractor to discard domain-specific information~\cite{DANN,DA_IMU}. Methods such as Deep Joint Distribution Optimal Transport (DeepJDOT)~\cite{DeepJDOT} further consider the relationship between feature and label distributions during alignment.

An alternative consists of performing adaptation at the image level. Instead of forcing the recognition network to directly learn domain-invariant representations, image translation methods transform labeled source samples so that their visual appearance approaches that of the target domain while their semantic content is preserved. This strategy is particularly relevant for handwritten documents because domain shifts frequently originate from visual characteristics such as writing style, stroke width, ink intensity, background texture, scanning conditions, and document degradation.

Disentangling Representation and Adaptation Networks (DRANet)~\cite{DRANet} follow this principle by decomposing image representations into content and style components. The content representation is intended to preserve the information shared between domains, whereas the style representation captures domain-dependent visual properties. By combining source content with target style, target-like images can be generated while retaining the semantic information associated with the labeled source samples. Such disentanglement provides an attractive mechanism for adapting recognition systems when target annotations are unavailable.

\subsection{HTR and Domain Adaptation}

To our knowledge, \cite{Imbert2026} and \cite{alfaro2024align} are the only works specifically addressing DA for HTR. While \cite{alfaro2024align} achieves insufficient recognition accuracy for practical use, \cite{Imbert2026} demonstrates substantially better performance. We therefore focus on \cite{Imbert2026}. The approach relies on three successive stages. First, a recognition model is trained on labeled source-domain data to learn an initial representation of the recognition task. Second, unsupervised DA is performed using labeled source data together with unlabeled target images. A disentanglement-based image translation model separates content and style information and transfers target-domain visual characteristics to source samples. This process produces novel images with original textual content from the source, but whose appearance aims to mimic that of the target domain. Finally, the HTR model is trained using these target-style labeled images, allowing target-domain visual information to be incorporated without requiring manual annotation of the target data.

For HTR, this strategy makes it possible to exploit unlabeled handwritten lines from a target collection while preserving the supervision provided by the source transcriptions. However, this approach presents two main limitations. 
First, the visual fidelity of the DA remains limited. Since HTR relies on fine-grained visual information, the generation of realistic target-style handwritten lines while accurately preserving character shapes, strokes, and textual content is particularly important. Improving the visual quality of the generated images could therefore provide more representative training samples for the target domain. 
Second, the pipeline focuses exclusively on visual adaptation and does not incorporate language modeling. Consequently, linguistic regularities and contextual dependencies between characters are not exploited during recognition. 
In this paper, we integrate post-HTR correction with visual DA to complement the visual predictions of the HTR model and improve recognition across domains with different visual and linguistic characteristics.

\subsection{Language modeling}

HTR models can implicitly learn linguistic information during training, particularly through autoregressive decoders~\cite{DAN}. However, this knowledge is tied to the training distribution and may generalize poorly under language shift. External language models can therefore be used to refine recognition predictions. N-gram models remain an effective approach, notably through lattice rescoring~\cite{meyer2026n}, while bidirectional masked language models such as BERT enable N-best hypothesis rescoring using pseudo-log-likelihood scores~\cite{shin2019effective,salazar2020masked}. More recently, VLMs and LLMs have been used to post-correct OCR predictions from text alone or jointly with the document image~\cite{10.1007/978-3-032-36042-7_14}. LLMs may hallucinate or alter the meaning of a text line if they are not carefully constrained \cite{boros-etal-2024-post}. These considerations motivate our choice of LLM-based post-correction, which avoids the need to train a dedicated LM for each dataset while offering the flexibility to incorporate the designed prompts and explicit correction rules. This approach provides a promising direction for improving HTR transcription accuracy while limiting undesired modifications to the recognized text.

\section{Method}\label{sec_method}

\begin{figure*}[!t]
\centering
\includegraphics[width=\linewidth]{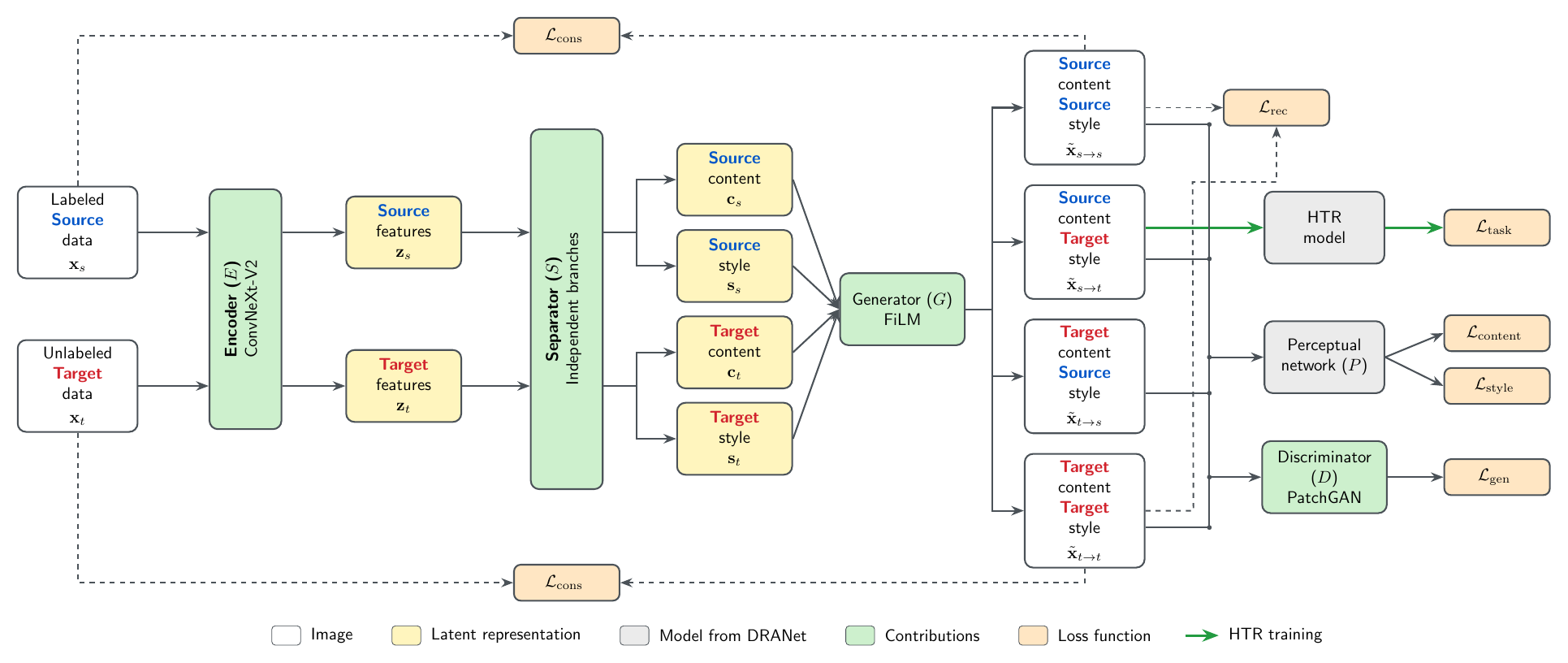}
\caption{Overview of the proposed DRA-HTR architecture. The overall content–style adaptation principle follows~\cite{Imbert2026}, while the architectures of its main components are redesigned in this work for HTR. The encoder ($E$) extracts visual features from source and target images, which are decomposed by the separator ($S$) into content and style representations. The generator ($G$) recombines source content with target style to generate target-style handwritten text lines while preserving their textual content. The generated images are constrained using adversarial, reconstruction, latent-consistency, and perceptual objectives. DA is performed only once during training to adapt the HTR model to the target domain. After adaptation, the model retains its original inference time.}
\label{fig_DRANET}
\end{figure*}

In this work, we introduce DRA-HTR, an extension of the previously proposed DA framework  introduced in~\cite{Imbert2026} by rethinking its main components for line-level HTR and incorporating post-HTR correction to further improve transcription accuracy. The proposed architecture is illustrated in Figure~\ref{fig_DRANET}.

The proposed framework retains the main principles of~\cite{Imbert2026}. An encoder extracts visual features from source and target images, while a separator disentangles these features into content and style representations. A generator then combines source content with target style to produce adapted images that preserve the original textual content while reproducing the visual characteristics of the target domain. A discriminator encourages the generation of visually realistic images, while perceptual constraints help maintain relevant visual characteristics.

Building upon this framework, we introduce two main improvements specifically designed for line-level HTR. First, we rethink the domain-adaptation components by selecting more suitable backbone architectures for the encoder, separator, generator, and discriminator. These components are highlighted in green in Figure~\ref{fig_DRANET}, while the unchanged HTR model and perceptual network are shown in gray. These architectural modifications aim to improve the visual realism of generated handwritten text lines while preserving their textual content. Second, we incorporate post-HTR correction through LLM-based post-correction to exploit linguistic dependencies and further improve transcription accuracy.

The proposed framework extends the original three-stage adaptation strategy to four stages. First, an HTR model is trained on labeled source-domain data. Second, unsupervised DA is performed using labeled source images and unlabeled target images to generate target-style handwritten text lines while preserving their original transcriptions. Third, the HTR model is trained using these generated images and their corresponding source transcriptions. Fourth, LLM-based post-correction  is applied to refine the recognition predictions and improve transcription accuracy. The following subsections describe the HTR model, the rethought domain-adaptation components, and the post-HTR correction approach in detail.

\subsection{HTR Model}\label{sec:Expert Model}

The HTR model constitutes the visual recognition component of the proposed framework. It is used both to perform HTR and to provide task-specific representations during DA. Unlike~\cite{Imbert2026}, which considered both isolated character classification and HTR, the present work focuses exclusively on line-level HTR.
The HTR architecture used in~\cite{Imbert2026} was originally introduced in~\cite{Retsinas22} and subsequently extended in~\cite{Corbille25}. In this work, we adopt the original CRNN architecture proposed by~\cite{Retsinas22} to reduce computational costs during training. 
Avoiding the additional computations of center loss class centers from~\cite{Corbille25} and the associated approximal optimization, the original architecture enables faster training, making it more suitable for our DA pipeline, which involves multiple training stages across several source–target domain pairs.
The architecture, follows a convolutional-recurrent design. Its convolutional backbone progressively extracts visual representations from an input handwritten text line through successive convolutional blocks with increasing numbers of feature channels. Squeeze-and-excitation modules~\cite{SELayer} are incorporated between these blocks to recalibrate feature channels and emphasize informative responses.

At the end of the convolutional backbone, column-wise max pooling converts the two-dimensional feature maps into a one-dimensional feature sequence while preserving the horizontal organization of the text line. This sequence is processed by three bidirectional LSTM layers with 256 hidden units to capture contextual dependencies in both reading directions. Finally, a linear layer maps the resulting representations to the output character classes.

The recognition network is trained using CTC loss, which enables the alignment between the input image and its transcription to be learned without explicit character-level segmentation. 

Within the proposed framework, this HTR model provides recognition-oriented information during adaptation and subsequently learns from the generated target-style images. Since the HTR model itself is inherited from the baseline \cite{Imbert2026}, it enables the isolation of the rethought DA pipeline's contributions, ensuring that improvements in recognition performance can be attributed primarily to the proposed adaptation strategy rather than to modifications of the HTR backbone.

\subsection{Domain Adaptation}\label{sub_DA}

The visual adaptation stage builds upon the DRANet \cite{DRANet} and the three-stage pipeline introduced in \cite{Imbert2026}. The fundamental principle of DRANet is to decompose an image representation into two complementary components; 
\begin{itemize}
    \item content: representing the textual information of an image, which must be preserved during style transfer to maintain consistency with its original transcription, even though textual content may differ across domains, 
    \item style: representing domain-dependent visual characteristics.
\end{itemize} 
For HTR, this decomposition is particularly useful because the textual sequence must remain unchanged while properties such as handwriting appearance, stroke width, contrast, background texture, and document degradation may vary substantially between datasets.

Let $\mathcal{D}_s = \{(\mathbf{x}_s, \mathbf{y}_s)\}$ denote the labeled source domain, where $\mathbf{x}_s$ is a text-line image and $\mathbf{y}_s$ its transcription, and let $\mathcal{D}_t = \{\mathbf{x}_t\}$ denote the unlabeled target domain.
Given a source image $\mathbf{x}_s$ and a target image $\mathbf{x}_t$, the adaptation network generates an image $\tilde{\mathbf{x}}_{s\rightarrow t}$ that preserves the textual content of $\mathbf{x}_s$ while reproducing visual characteristics extracted from $\mathbf{x}_t$. Because the textual sequence is preserved, the $\mathbf{y}_s$ can be directly assigned to $\tilde{\mathbf{x}}_{s\rightarrow t}$. The generated images therefore constitute labeled target-style training samples without requiring target-domain transcriptions.

As in~\cite{Imbert2026}, the complete training procedure contains three stages. First, the HTR model is pretrained on labeled source-domain data. Second, labeled source images and unlabeled target images are used to optimize the domain-adaptation model. Third, source images translated toward the target domain are paired with their original source transcriptions and used to train the HTR model.

The main difference from~\cite{Imbert2026} lies in the architecture used to perform the second stage. The previous framework employed a comparatively generic DRANet configuration and demonstrated that content–style adaptation could benefit HTR. In this work, we go beyond applying this generic framework to handwritten lines and redesign its principal components specifically for HTR. The new architecture targets two limitations that are particularly important for line recognition: preserving fine character and stroke structures during translation and producing visually realistic samples over long and spatially heterogeneous text lines. The novel components introduced in this work are described next.

\begin{itemize}

\item \textbf{Encoder ($E$):}
A major architectural change with respect to~\cite{Imbert2026} is the replacement of the previous feature extractor, which consisted of the convolutional layers of the HTR model, with a ConvNeXt-V2 backbone~\cite{ConvNeXtV2}. For an image $\mathbf{x}_d$ from domain $d\in\{s,t\}$, the encoder produces
\[
    \mathbf{z}_d = E(\mathbf{x}_d).
\]
The ConvNeXt-V2 encoder is initialized from large-scale image pretraining, providing generic visual representations before DA training. Its hierarchical architecture captures information at multiple spatial scales: local representations describe fine structures such as strokes, edges, and texture, whereas deeper representations encode larger spatial configurations across the handwritten line.
This multiscale representation is particularly relevant to the content-style decomposition required by HTR. The adaptation network must alter visual properties associated with the target collection without modifying character shapes or their spatial ordering. A stronger hierarchical representation therefore provides the separator with richer information from which recognition-relevant structure and domain-dependent appearance can be disentangled.

\item \textbf{Separator ($S$):}
In the separator used by~\cite{Imbert2026}, spectrally normalized convolutional layers produce a spatial style map from the encoded feature $\mathbf{z}_d$. The content representation is then obtained by subtracting this style map from the encoded features:
\[
\mathbf{s}_d=S(\mathbf{z}_d),
\qquad
\mathbf{c}_d=\mathbf{z}_d-\mathbf{s}_d.
\]
During source-to-target adaptation, a learned weighting tensor specific to each domain conversion is applied elementwise to the source content:
\[
\mathbf{c}_{s\rightarrow t
=\mathbf{w}_{s\rightarrow t}\odot\mathbf{c}_s}.
\]
These weights vary across channels and spatial positions and are shared across samples for a given conversion. They are initialized to one and are not constrained to $[0,1]$, allowing them to attenuate, amplify, or reverse the sign of content features.
However, this residual decomposition introduces an algebraic ambiguity:
\[
\mathbf{c}_d+\mathbf{s}_d=\mathbf{z}_d
\]
holds regardless of the predicted style map. This identity does not impose any constraint on how information is distributed between content and style. In particular, a zero style map and a content representation equal to the encoded features satisfy the decomposition.
To remove this algebraic ambiguity, we replace the residual decomposition with two separately parameterized branches:
\[
\mathbf{c}_d=f_{\mathrm{content}}(\mathbf{z}_d),
\qquad
\mathbf{s}_d=f_{\mathrm{style}}(\mathbf{z}_d).
\]
The content branch uses two spectrally normalized convolutional layers, each followed by instance normalization and ReLU activation, to produce a spatial content representation. In contrast, the style branch uses a spectrally normalized convolution followed by ReLU activation, global average pooling, a spectrally normalized linear projection layer, layer normalization, and Tanh activation to extract a compact style vector.

During source-to-target adaptation, the separator preserves the source content without conversion-specific weighting:
\[
\mathbf{c}_{s\rightarrow t}=\mathbf{c}_s.
\]
The separator thus provides a spatial source content representation and a compact target style vector for subsequent image synthesis. Unlike the residual decomposition, the two branches are not constrained to sum to the encoded features. Global average pooling removes the spatial layout from the style representation, while the content branch retains spatial structure. These architectural choices encourage complementary representations, although meaningful content--style disentanglement still depends on the training objectives and optimization.

\item \textbf{Generator ($G$):}
The generator is redesigned to reconstruct handwritten text lines from the disentangled representations. Source-to-target adaptation is performed by combining adapted source content with target style:
\[
    \tilde{\mathbf{x}}_{s\rightarrow t}
    =
    G\left(\mathbf{c}_{s\rightarrow t}+\mathbf{s}_t\right).
\]
Unlike the generator employed in~\cite{Imbert2026}, our architecture focuses on progressive spatial reconstruction and the preservation of fine handwriting details. The separator provides a spatial content representation from the source image and a style vector from a target-domain image. To incorporate target style information throughout the decoding process, we employ Feature-wise Linear Modulation (FiLM)~\cite{Perez2018FiLM}, which allows the target style vector to modulate the source content features at each decoder stage. This mechanism enables the target style to control how the source content is visually reconstructed.

Formally, let \(\mathbf{c}_s\) denote the spatial content representation of the source image and \(\mathbf{s}_t\) the style vector of a target-domain image. The content representation initializes the decoding process and provides the spatial structure of the source text. At decoder stage \(k\), the intermediate feature map derived from \(\mathbf{c}_s\) is modulated according to
\[
\mathbf{h}_k=
\operatorname{ReLU}\!\left(
\left(1+\gamma_k(\mathbf{s}_t)\right)
\odot F_k(\mathbf{c}_s)
+\beta_k(\mathbf{s}_t)
\right),
\]
where $F_k(\mathbf{c}_s)$ denotes the instance-normalized decoder features, while $\gamma_k(\mathbf{s}_t)$ and $\beta_k(\mathbf{s}_t)$ are learned channel-wise scaling offsets and shifts derived from the target style vector. The operator $\odot$ denotes channel-wise multiplication, and the term $1+\gamma_k(\mathbf{s}_t)$ enables modulation around unit scaling.
Through this mechanism, the source content determines the spatial structure and character sequence, while the target style controls their visual appearance. Applying FiLM at successive decoder stages enables the generator to progressively incorporate target-domain handwriting characteristics while aiming to preserve the source text. This distinction is important for HTR, where small reconstruction artifacts may modify characters, merge strokes, or remove diacritics, making the original transcription invalid. Finally, the generated images retain the grayscale format, normalized dimensions, and intensity range of the original HTR inputs, allowing their direct use for subsequent HTR model training.

\item \textbf{Discriminator ($D$):}
In the baseline adaptation method, the discriminator is comprised of seven convolutional layers with ReLU activations, followed by global average pooling and a fully connected layer with sigmoid activation to produce a single \mbox{real/fake} probability per image. In contrast, the proposed configuration employs a PatchGAN discriminator, which produces a spatial grid of \mbox{real/fake} scores corresponding to local image regions. Rather than assigning a single \mbox{real/fake} prediction to an entire text-line image, the discriminator evaluates multiple spatial regions.
This formulation is appropriate for HTR because many domain-dependent characteristics occur locally: stroke thickness, ink texture, background degradation, contrast, scanning artifacts, and acquisition noise. At the same time, handwritten lines may exhibit highly variable widths and textual layouts, making a single global realism criterion less appropriate.
The discriminator is trained to distinguish patches extracted from real target-domain images from patches of generated source-to-target images, whereas the generator is optimized to make the translated patches indistinguishable from target-domain observations. Spectral normalization is employed to improve adversarial training stability.
\item \textbf{Perceptual Network ($P$):}
Following the original DRANet formulation, the perceptual network remains a frozen VGG-19 model pretrained on ImageNet. Features extracted at different representation levels are used to preserve the semantic and spatial organization of the source text line while encouraging the generated image to reproduce the appearance of the target domain.
The modification introduced in our approach targets the complementary reconstruction objectives rather than the perceptual network itself. In addition to the conventional reconstruction and latent-consistency losses, we employ foreground-sensitive reconstruction and image-gradient constraints to emphasize handwritten strokes. These constraints reduce excessive smoothing, missing strokes, and local structural distortions that could produce visually plausible images while modifying their transcription.
\end{itemize}

The complete adaptation objective combines adversarial, reconstruction, perceptual-content, perceptual-style, and latent-consistency terms, together with the proposed foreground- and gradient-sensitive constraints. These objectives address complementary aspects of the translation problem: adversarial learning promotes target-domain realism, perceptual and style constraints reproduce target appearance, while reconstruction, latent consistency, foreground weighting, and gradient preservation constrain the transformation to maintain the textual structure inherited from the source image.

The three-stage training principle and the use of unlabeled target data are retained, while the domain-adaptation architecture is substantially redesigned to improve the visual fidelity and transcription preservation of generated handwritten lines. The resulting target-style labeled samples are subsequently used to train the HTR model. 
Finally, post-HTR correction complements this visual adaptation by incorporating linguistic information during recognition, as described in the following subsection.

\subsection{Post-HTR correction}
\label{sec:Language modeling}

In addition to DRA-HTR, we introduce DRAL-HTR, which integrates Qwen3.8-27B-based post-correction to improve HTR transcription accuracy. To limit hallucinations and guide the correction process, we provide multiple recognition hypotheses, neighboring lines as contextual information, and explicit correction rules. The approach uses the HTR candidates to correct plausible recognition errors while preserving historical spelling and the original meaning of the text.

For each text line, a CTC prefix beam search with a beam width of 150 is performed, and the 20 highest-scoring distinct hypotheses, ranked by their CTC log-probabilities, are retained. These hypotheses are then passed to Qwen, which is instructed to select or combine plausible recognition hypotheses, allowing only minimal modifications beyond them to correct likely character, word-boundary, or punctuation errors. Neighboring lines provide contextual information but must not be included in the output. Explicit rules prevent paraphrasing, translation, spelling modernization, and unsupported additions, while preserving historical spelling, names, abbreviations, and the original meaning. The model outputs only the corrected transcription, without confidence values. The complete prompt is provided in Listing~\mbox{\ref{lst:qwen_post_correction_prompt}}.

\begin{lstlisting}[
    float=!tp,
    language=Python,
    caption={Prompt used for HTR post-correction with Qwen.},
    label={lst:qwen_post_correction_prompt},
    basicstyle=\ttfamily\fontsize{7}{7}\selectfont,
    breaklines=true,
    breakatwhitespace=true,
    columns=fullflexible,
    keepspaces=true,
    showstringspaces=false,
    frame=single
]
You post-correct one line of handwriting-recognition (HTR) output. The expected language and domain are {language_context}.
The input may contain these sections:
- CANDIDATES: the recognizer's best readings of the same line image, ranked by the recognizer's relative confidence. The correct text is often one of them, or a combination of parts of several of them, and it is frequently not the first one.
- PREVIOUS LINE / NEXT LINE: recognized neighbouring lines of the same page. They may also contain errors. Use them only to understand the topic, names, and spelling conventions; never copy their text into the output.
Follow these rules:
1. Choose the candidate that is the most plausible text in the expected language, or combine parts of different candidates when each is correct in a different place. The ranking and percentages are only a weak hint: a lower-ranked candidate made of real words beats a higher-ranked one that is not.
2. Go beyond the candidates only for plausible recognition errors:
   - character substitutions, insertions, or deletions
   - incorrect word boundaries (splits or merges)
   - punctuation recognition errors
3. Use the expected language/domain only to choose between plausible readings of the recognized text.
4. Preserve historical spelling, archaic forms, dialect, names, foreign words or phrases, abbreviations, numbers, and original meaning. Keep capitalization unless the candidates disagree on it or it is clearly a recognition error.
5. Do not modernize spelling or grammar.
6. Do not translate, paraphrase, explain, normalize, or stylistically improve the text.
7. Do not add words based on what you think the surrounding context should say.
8. Only when candidates are equally plausible, prefer the higher-ranked one.
9. When going beyond the candidates, make the minimum number of edits necessary.
10. Return only the corrected line in the corrected_text field, without numbering or confidence values.
\end{lstlisting}

\subsection{Training and Optimization Strategies}\label{sub_training_tricks}

The joint optimization of the adaptation and recognition components is particularly sensitive to unstable adversarial gradients and to the quality of the images generated during the early training iterations.
We therefore introduce several training strategies to improve convergence and preserve the textual content of the adapted images.

\begin{itemize}

\item \textbf{Progressive introduction of recognition supervision.}
The recognition objective is not introduced at the beginning of adaptation. The encoder, separator, generator, and discriminator are first allowed to learn reconstruction and style-transfer functions without being constrained by recognition predictions obtained from initially low-quality generated images. Recognition-related optimization is activated only after a predefined number of iterations.
This delayed schedule reduces the risk that unreliably generated samples destabilize the recognition model or that noisy recognition gradients constrain the generator before it has learned an appropriate target-domain style.
\item \textbf{Recognition-guided image generation.}
Once the initial adaptation stage is completed, a CTC-based recognition loss is added to the objective used to optimize the encoder, separator, and generator. For a generated source-to-target image $\tilde{\mathbf{x}}_{s\rightarrow t}$ with source transcription $\mathbf{y}_s$, and denoting the HTR model by $T$, this constraint is expressed as
\begin{equation}
    \mathcal{L}_{\mathrm{task}}
    =
    \mathcal{L}_{\mathrm{CTC}}\!\left(
    T(\tilde{\mathbf{x}}_{s\rightarrow t}),\mathbf{y}_s
    \right).
\end{equation}
During this adaptation update, the parameters of both the recognition model and the discriminator are frozen. Consequently, the CTC gradient is propagated through the recognition model to the generated image, but only the adaptation components are updated. The recognition model therefore acts as a fixed semantic constraint that penalizes transformations that modify the transcription.
The complete adaptation objective becomes
\begin{equation}
\begin{aligned}
    \mathcal{L}_{\mathrm{adapt}}
    ={}&
    \lambda_{\mathrm{gen}}\mathcal{L}_{\mathrm{gen}}
    \\
    &+ \lambda_{\mathrm{rec}}\mathcal{L}_{\mathrm{rec}}
    \\
    &+ \lambda_{\mathrm{content}}\mathcal{L}_{\mathrm{content}}
    \\
    &+ \lambda_{\mathrm{style}}\mathcal{L}_{\mathrm{style}}
    \\
    &+ \lambda_{\mathrm{cons}}\mathcal{L}_{\mathrm{cons}} \\
    &+ \lambda_{\mathrm{task}}\mathcal{L}_{\mathrm{task} .} 
\end{aligned}
\end{equation}

\begin{figure*}[b]
     \centering
     \includegraphics[width=\textwidth]{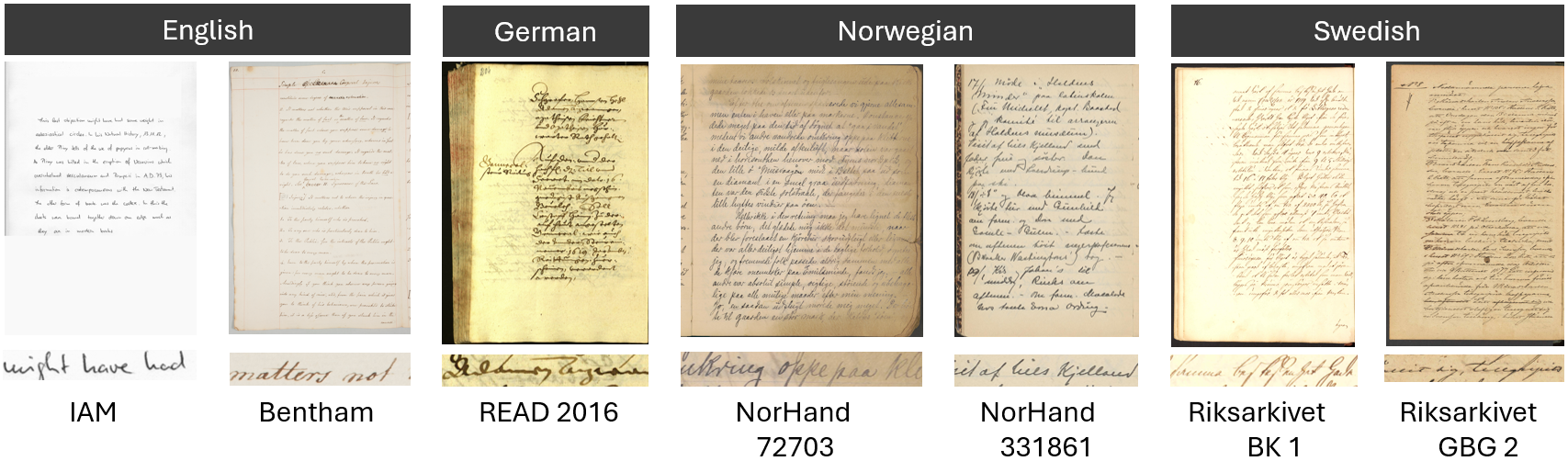}
     \caption{Examples of handwritten document images from the datasets used in our experiments. The examples illustrate the diversity of handwriting styles, document layouts, and paper appearance across the different datasets. A cropped text line is shown below each document image to show the writing style.}
     \label{fig:datasets}
\end{figure*} 


Here, $\mathcal{L}_{\mathrm{gen}}$ denotes the adversarial generator loss, which encourages the generation of realistic target-style images; $\mathcal{L}_{\mathrm{rec}}$ is the reconstruction loss, which promotes the reconstruction of the original images; $\mathcal{L}_{\mathrm{content}}$ and $\mathcal{L}_{\mathrm{style}}$ enforce the preservation of content and consistency of style representations, respectively; $\mathcal{L}_{\mathrm{cons}}$ is the consistency loss, which encourages coherence between the original and reconstructed latent representations; and $\mathcal{L}_{\mathrm{task}}$ is the CTC-based recognition loss, which ensures that the generated images remain consistent with their source transcriptions.
The coefficients $\lambda_{\mathrm{gen}}$, $\lambda_{\mathrm{rec}}$, $\lambda_{\mathrm{content}}$, $\lambda_{\mathrm{style}}$, $\lambda_{\mathrm{cons}}$, and $\lambda_{\mathrm{task}}$ are fixed, empirically chosen weighting coefficients that balance the contributions of the different objectives during training.
This task-guided formulation encourages the generator to produce images that are both visually consistent with the target domain and recognizable under the original source transcription.

\item \textbf{Gradient Accumulation.}
Since larger batch sizes are important for stable DA training, we use gradient accumulation to achieve an effectively larger batch size while respecting GPU-memory constraints.
\end{itemize}

\section{Experiments}\label{sec_expe}

This section describes the datasets, preprocessing steps, and experimental protocol used to evaluate the proposed domain-adaptation pipeline. We consider handwritten document collections covering four languages: English, German, Norwegian, Swedish. This diversity provides a challenging experimental setting for evaluating the ability of the proposed approach to transfer knowledge from labeled source data to unlabeled target domains.

\subsection{Datasets}\label{subsec:datasets}

We evaluate the proposed approach on 7 handwritten datasets covering four languages. The selected collections include both modern and historical documents and exhibit substantial variations in handwriting style, document age, layout, paper appearance, and acquisition conditions, providing a diverse setting for evaluating DA. For each language, the different datasets or document collections are treated as distinct domains, allowing adaptation to be performed between datasets sharing the same language. Representative document examples are shown in Fig.~\ref{fig:datasets}, while Table~\ref{tab:datasets} reports the number of pages and text lines in the training, validation, and test sets.

The English datasets comprise IAM~\cite{Marti2002IAM}, containing modern handwritten documents, and Bentham~\cite{Sanchez2014Bentham}, consisting of historical manuscripts. For Norwegian, we use the NorHand 72702 and NorHand 331861 partitions from~\cite{pathirage2026generalization}, corresponding to the two largest books in the NorHand v3 collection~\cite{NorHandv3}. Similarly, for Swedish, we use the Riksarkivet BK 1 and Riksarkivet GBG 2 partitions from~\cite{pathirage2026generalization}. These four partitions provide training, validation, and test sets of comparable sizes, enabling controlled cross-domain evaluation. Finally, the German dataset READ 2016~\cite{Sanchez2016READ} comprises historical handwritten documents.

\begin{table*}[t]
\centering
\footnotesize
\caption{Datasets used in the experiments. The number of pages and text lines is reported for each training, validation, and test set.}
\label{tab:datasets}
\resizebox{\textwidth}{!}{
\begin{tabular}{llrrrrrrr}
\toprule
& & \multicolumn{2}{c}{\textbf{Train}} &
\multicolumn{2}{c}{\textbf{Validation}} &
\multicolumn{2}{c}{\textbf{Test}} &
\textbf{Total Lines} \\
\cmidrule(lr){3-4}
\cmidrule(lr){5-6}
\cmidrule(lr){7-8}
\textbf{Language} & \textbf{Dataset} &
\textbf{Pages} & \textbf{Lines} &
\textbf{Pages} & \textbf{Lines} &
\textbf{Pages} & \textbf{Lines} & \\
\midrule

English
& IAM
& 747 & 6,482 & 116 & 976 & 336 & 2,915 & 10,373 \\
& Bentham
& 350 & 9,198 & 50 & 1,415 & 33 & 860 & 11,473 \\

\midrule

German
& READ 2016
& 350 & 8,367 & 50 & 1,043 & 50 & 1,140 & 10,550 \\

\midrule

Norwegian
& NorHand 72702
& 292 & 6,069 & 37 & 792 & 37 & 753 & 7,614 \\
& NorHand 331861
& 319 & 6,053 & 40 & 784 & 40 & 766 & 7,603 \\

\midrule

Swedish
& Riksarkivet BK 1
& 184 & 8,824 & 23 & 1,147 & 24 & 1,117 & 11,088 \\
& Riksarkivet GBG 2
& 146 & 10,954 & 18 & 1,319 & 19 & 1,436 & 13,709 \\

\bottomrule
\end{tabular}}
\end{table*}

\subsection{Data Preprocessing}
\label{subsec:preprocessing}

Each dataset contains a distinct character set, resulting in different numbers of output classes for the HTR models. To ensure consistency across datasets, we standardize the character labels. We also exclude rare or unusual characters and those with insufficient occurrences for reliable training ($\leq 25$) . Since our DA framework requires a shared character set across all datasets, we construct a unified vocabulary by taking the union of the standardized character sets, resulting in 87 output classes.

For all datasets except IAM, we extract text-line images using polygon coordinates. For IAM, we retain the original line images, as the handwritten text is already surrounded by a clean white background. The resulting images are then converted to grayscale.

\subsection{Training and Evaluation}
\label{subsec:training_evaluation}

The experiments follow the four-stage pipeline presented in Sec.~\ref{sec_method}. First, the HTR model is trained on the labeled source-domain data. Second, the DA model is trained using labeled source images together with unlabeled images from the target domain. The resulting model is then used to translate source-domain text lines into the visual style of the target domain while preserving their textual content and associated transcriptions. Finally, the HTR model is trained using the generated target-style images and their inherited source labels. Target-domain transcriptions are never used during the adaptation process.

\subsubsection{HTR model Training}

The CRNN model is trained on an NVIDIA A100 40GB GPU with a batch size of 32, using input images resized to a height of 160 pixels and a maximum width of 1570 pixels within the pipeline. The Adam optimizer uses an initial learning rate of $1 \times 10^{-4}$, which is reduced by a factor of 10 with a patience of 100 epochs. Following this reduction, early stopping is applied with a patience of 200 epochs. The same training procedure is used for both source-only training and fine-tuning on generated target-style data to ensure a fair comparison.

\subsubsection{Domain Adaptation Training}
The DA framework is trained on an NVIDIA A100 40GB GPU with a mini-batch size of 4. Since effective domain alignment requires sufficiently diverse source and target samples within each optimization step, gradient accumulation is employed to overcome GPU memory limitations. Gradients are accumulated over 8 consecutive mini-batches, resulting in an effective batch size of 32. This provides a more representative sampling of both domains during optimization without increasing the memory required for each forward and backward pass.

The adaptation framework is optimized using the Adam optimizer. The learning rates of the different components and the weights associated with the adaptation losses are summarized in Table~\ref{tab:hyperparameters}. These hyperparameters control the balance between visual adaptation and preservation of the textual content in the generated images.

\begin{table}[h]
\centering
\caption{Hyperparameters used for training the DA framework.}
\label{tab:hyperparameters}
\begin{tabular}{ll}
\toprule
\textbf{Parameter} & \textbf{Value} \\
\midrule
\multicolumn{2}{l}{\textit{Learning rates}} \\
Generative components       & $2 \times 10^{-4}$ \\
HTR model                   & $4 \times 10^{-4}$ \\
Separator                   & $2 \times 10^{-4}$ \\
\midrule
\multicolumn{2}{l}{\textit{Loss weights}} \\
Generator loss             & 0.1 \\
Content preservation loss    & $2 \times 10^{-4}$ \\
Style loss                   & 1.0 \\
Consistency loss             & 0.12 \\
Reconstruction loss          & 0.2 \\
Recognition loss             & 0.5 \\
\bottomrule
\end{tabular}
\end{table}

\subsubsection{Evaluation}
Recognition performance is evaluated using the standard Character Error Rate (CER). For each source-target configuration, the complete experiment is repeated 3 times, and the mean and standard deviation are reported. For each run, model selection is performed on the validation set according to the CER, and the checkpoint achieving the lowest validation CER is subsequently evaluated on the corresponding test set.

We evaluate the different stages of the proposed approach under the same experimental conditions. The source-only configuration measures the recognition performance obtained when the HTR model is trained exclusively on labeled source-domain data. The proposed DA configuration evaluates the benefit of training the recognizer with source samples adaptation into the target-domain style. Finally, language modeling is incorporated during decoding to evaluate the complementary contribution of linguistic information.

\subsection{Results}

This section evaluates the proposed DA pipeline across seven HTR datasets. We first compare our HTR models with SOTA results and assess their cross-dataset generalization. We then evaluate the effectiveness of our DA method against models without adaptation. Finally, an ablation study examines improvements over~\cite{Imbert2026}, the impact of architectural modifications, and the contribution of post-HTR correction.

\subsubsection{SOTA on Each Dataset}

To justify our choice of HTR model, we compare its performance with the SOTA on each dataset. To this end, we train the HTR model architecture described in Section~\ref{sec:Expert Model} separately on each of the seven datasets in a supervised manner.

Table~\ref{tab:sota_results} compares the results obtained by our HTR models with the corresponding SOTA results on each dataset. Our HTR models outperform the SOTA results on four of the seven datasets while achieving competitive performance on the remaining three.

\begin{table*}[t]
\centering
\caption{Comparison of SOTA HTR results with our HTR models on each dataset. Recognition is evaluated on the test set with Character Error Rate (CER) ($\downarrow$)}
\label{tab:sota_results}
\resizebox{\textwidth}{!}{
\begin{tabular}{lccccccc}
\toprule
\multirow{2}{*}{\textbf{Dataset}} &
\multirow{2}{*}{\textbf{IAM}} &
\multirow{2}{*}{\textbf{Bentham}} &
\multirow{2}{*}{\textbf{READ 2016}} &
\textbf{NorHand} &
\textbf{NorHand} &
\textbf{Riksarkivet} &
\textbf{Riksarkivet} \\

{} & {} & {} & {} &
\textbf{72702} &
\textbf{331861} &
\textbf{BK 1} &
\textbf{GBG 2} \\
\midrule

\textbf{SOTA CER (\%)} &
\textbf{3.36} \cite{Gu_2025_CVPR} &
2.37 \cite{Chinthaginjala2024} &
3.80 \cite{COQUENET2026113373} &
1.61 \cite{pathirage2026generalization} &
\textbf{3.09} \cite{pathirage2026generalization} &
\textbf{3.40}  \cite{pathirage2026generalization}&
2.96 \cite{pathirage2026generalization} \\

\textbf{Our HTR model} &
3.83 $\pm$\,{\small 0.15} &
\textbf{2.12} $\pm$\,{\small 0.10} &
\textbf{3.55} $\pm$\,{\small 0.02} &
\textbf{1.54} $\pm$\,{\small 0.07} &
3.65 $\pm$\,{\small 0.15} &
3.55 $\pm$\,{\small 0.26} &
\textbf{2.58} $\pm$\,{\small 0.18} \\
\bottomrule
\end{tabular}}
\end{table*}

These results establish a strong recognition baseline, demonstrating that the selected CRNN based HTR model is suitable for integration into the DA pipeline.

\subsubsection{HTR Model on each dataset}

Having established the effectiveness of our HTR model, we now investigate its ability to generalize to unseen datasets and assess the need for DA. To this end, we evaluate each of our seven HTR models trained in Table~\ref{tab:sota_results} on the six remaining datasets in a zero-shot setting. The results are presented in Table~\ref{tab:cross_dataset_comparison}(a).

The results reveal a significant lack of generalization across datasets, with CER values exceeding 30\% in nearly all cross-dataset evaluations, even when the source and target datasets share the same language. Only two exceptions are observed: adaptation from IAM to Bentham, achieving a CER of 13.23\%, and from Nor33186 to Nor7770, achieving a CER of 17.56\%.

These findings demonstrate that, despite achieving strong recognition performance on their respective source datasets, the HTR models struggle to generalize to unseen handwriting styles and documents. This highlights the limitations of zero-shot cross-dataset recognition and motivates the introduction of DA methods to improve recognition performance on unseen datasets.

\begin{table*}[t]
\centering
\caption{Comparison of CER (\%) across seven HTR datasets.
Rows indicate the source training dataset and columns indicate
the target evaluation dataset.}
\label{tab:cross_dataset_comparison}

\resizebox{\textwidth}{!}{%
\begin{tabular}{l*{7}{r@{\,}l}}
\toprule
\multirow{2}{*}{\textbf{Source $\backslash$ Target}}
& \multicolumn{2}{c}{\multirow{2}{*}{\textbf{IAM}}}
& \multicolumn{2}{c}{\multirow{2}{*}{\textbf{Bentham}}}
& \multicolumn{2}{c}{\multirow{2}{*}{\textbf{READ 2016}}}
& \multicolumn{2}{c}{\textbf{NorHand}}
& \multicolumn{2}{c}{\textbf{NorHand}}
& \multicolumn{2}{c}{\textbf{Riksarkivet}}
& \multicolumn{2}{c}{\textbf{Riksarkivet}} \\

& \multicolumn{2}{c}{}
& \multicolumn{2}{c}{}
& \multicolumn{2}{c}{}
& \multicolumn{2}{c}{\textbf{72702}}
& \multicolumn{2}{c}{\textbf{331861}}
& \multicolumn{2}{c}{\textbf{BK 1}}
& \multicolumn{2}{c}{\textbf{GBG 2}} \\
\midrule


\multicolumn{15}{l}{\textbf{(a) HTR model}} \\
\midrule

\textbf{IAM}
& 3.83 & $\pm$\,{\small 0.15}
& 13.23 & $\pm$\,{\small 0.60}
& 77.48 & $\pm$\,{\small 1.42}
& 36.99 & $\pm$\,{\small 1.41}
& 55.35 & $\pm$\,{\small 0.88}
& 62.39 & $\pm$\,{\small 0.59}
& 55.96 & $\pm$\,{\small 0.69} \\

\textbf{Bentham}
& 29.23 & $\pm$\,{\small 0.85}
& 2.12 & $\pm$\,{\small 0.10}
& 74.09 & $\pm$\,{\small 0.90}
& 33.17 & $\pm$\,{\small 0.33}
& 59.48 & $\pm$\,{\small 0.24}
& 61.01 & $\pm$\,{\small 0.56}
& 49.88 & $\pm$\,{\small 0.45} \\

\textbf{READ 2016}
& 82.95 & $\pm$\,{\small 1.91}
& 69.35 & $\pm$\,{\small 3.35}
& 3.55 & $\pm$\,{\small 0.02}
& 61.70 & $\pm$\,{\small 2.84}
& 70.65 & $\pm$\,{\small 0.98}
& 60.57 & $\pm$\,{\small 1.30}
& 75.58 & $\pm$\,{\small 2.43} \\

\textbf{NorHand 72702}
& 64.60 & $\pm$\,{\small 2.71}
& 33.54 & $\pm$\,{\small 0.92}
& 71.61 & $\pm$\,{\small 0.36}
& 1.54 & $\pm$\,{\small 0.07}
& 36.78 & $\pm$\,{\small 0.61}
& 50.00 & $\pm$\,{\small 0.31}
& 44.08 & $\pm$\,{\small 0.33} \\

\textbf{NorHand 331861}
& 73.69 & $\pm$\,{\small 3.25}
& 48.24 & $\pm$\,{\small 1.21}
& 75.90 & $\pm$\,{\small 0.71}
& 17.56 & $\pm$\,{\small 0.67}
& 3.65 & $\pm$\,{\small 0.15}
& 57.18 & $\pm$\,{\small 0.32}
& 49.77 & $\pm$\,{\small 0.57} \\

\textbf{Riksarkivet BK 1}
& 58.56 & $\pm$\,{\small 0.46}
& 38.06 & $\pm$\,{\small 2.45}
& 63.97 & $\pm$\,{\small 0.32}
& 42.04 & $\pm$\,{\small 1.42}
& 59.90 & $\pm$\,{\small 0.82}
& 3.55 & $\pm$\,{\small 0.26}
& 49.83 & $\pm$\,{\small 1.46} \\

\textbf{Riksarkivet GBG 2}
& 75.03 & $\pm$\,{\small 4.32}
& 36.91 & $\pm$\,{\small 1.50}
& 74.44 & $\pm$\,{\small 0.32}
& 32.32 & $\pm$\,{\small 0.64}
& 53.03 & $\pm$\,{\small 1.10}
& 48.20 & $\pm$\,{\small 0.93}
& 2.58 & $\pm$\,{\small 0.18} \\

\midrule


\multicolumn{15}{l}{\textbf{(b) DRA-HTR }} \\
\midrule

\textbf{IAM}
& \multicolumn{2}{c}{--}
& 5.29 & $\pm$\,{\small 0.16}
& 16.59 & $\pm$\,{\small 1.27}
& 5.09 & $\pm$\,{\small 0.16}
& 16.04 & $\pm$\,{\small 0.55}
& 18.77 & $\pm$\,{\small 0.44}
& 19.29 & $\pm$\,{\small 0.19} \\

\textbf{Bentham}
& 9.94 & $\pm$\,{\small 0.21}
& \multicolumn{2}{c}{--}
& 20.85 & $\pm$\,{\small 0.75}
& 5.36 & $\pm$\,{\small 0.12}
& 17.82 & $\pm$\,{\small 0.27}
& 17.17 & $\pm$\,{\small 0.22}
& 22.05 & $\pm$\,{\small 5.29} \\

\textbf{READ 2016}
& 18.08 & $\pm$\,{\small 0.38}
& 13.53 & $\pm$\,{\small 0.40}
& \multicolumn{2}{c}{--}
& 7.72 & $\pm$\,{\small 0.36}
& 18.39 & $\pm$\,{\small 0.47}
& 16.52 & $\pm$\,{\small 0.49}
& 22.79 & $\pm$\,{\small 0.76} \\

\textbf{NorHand 72702}
& 13.89 & $\pm$\,{\small 0.36}
& 11.60 & $\pm$\,{\small 3.28}
& 16.86 & $\pm$\,{\small 0.73}
& \multicolumn{2}{c}{--}
& 10.93 & $\pm$\,{\small 0.23}
& 15.41 & $\pm$\,{\small 0.49}
& 16.35 & $\pm$\,{\small 0.60} \\

\textbf{NorHand 331861}
& 17.13 & $\pm$\,{\small 0.65}
& 12.25 & $\pm$\,{\small 0.12}
& 17.58 & $\pm$\,{\small 1.55}
& 4.27 & $\pm$\,{\small 0.06}
& \multicolumn{2}{c}{--}
& 19.03 & $\pm$\,{\small 0.31}
& 17.62 & $\pm$\,{\small 0.34} \\

\textbf{Riksarkivet BK 1}
& 20.04 & $\pm$\,{\small 0.21}
& 11.54 & $\pm$\,{\small 0.18}
& 19.90 & $\pm$\,{\small 0.40}
& 7.22 & $\pm$\,{\small 0.09}
& 21.93 & $\pm$\,{\small 0.61}
& \multicolumn{2}{c}{--}
& 18.35 & $\pm$\,{\small 0.95} \\

\textbf{Riksarkivet GBG 2}
& 30.03 & $\pm$\,{\small 8.20}
& 12.93 & $\pm$\,{\small 4.52}
& 18.60 & $\pm$\,{\small 0.16}
& 9.07 & $\pm$\,{\small 2.69}
& 14.32 & $\pm$\,{\small 0.69}
& 18.17 & $\pm$\,{\small 6.88}
& \multicolumn{2}{c}{--} \\

\midrule


\multicolumn{15}{l}{\textbf{(c) DRANet applied to HTR ~\cite{Imbert2026}}} \\
\midrule

\textbf{IAM}
& \multicolumn{2}{c}{--}
& 8.72 & $\pm$\,{\small 2.40}
& 20.61 & $\pm$\,{\small 4.65}
& 14.22 & $\pm$\,{\small 9.54}
& 18.65 & $\pm$\,{\small 1.80}
& 26.41 & $\pm$\,{\small 1.72}
& 30.07 & $\pm$\,{\small 6.86} \\

\textbf{Bentham}
& 13.53 & $\pm$\,{\small 3.18}
& \multicolumn{2}{c}{--}
& 25.66 & $\pm$\,{\small 3.96}
& 12.82 & $\pm$\,{\small 4.42}
& 27.21 & $\pm$\,{\small 2.12}
& 30.34 & $\pm$\,{\small 5.71}
& 21.74 & $\pm$\,{\small 0.76} \\

\textbf{READ 2016}
& 27.20 & $\pm$\,{\small 4.80}
& 22.52 & $\pm$\,{\small 1.00}
& \multicolumn{2}{c}{--}
& 23.26 & $\pm$\,{\small 16.33}
& 28.30 & $\pm$\,{\small 5.31}
& 27.68 & $\pm$\,{\small 7.87}
& 59.98 & $\pm$\,{\small 25.26} \\

\textbf{NorHand 72702}
& 21.79 & $\pm$\,{\small 9.33}
& 20.87 & $\pm$\,{\small 5.55}
& 22.12 & $\pm$\,{\small 7.20}
& \multicolumn{2}{c}{--}
& 19.14 & $\pm$\,{\small 4.57}
& 36.54 & $\pm$\,{\small 8.72}
& 26.05 & $\pm$\,{\small 6.82} \\

\textbf{NorHand 331861}
& 18.99 & $\pm$\,{\small 1.04}
& 15.84 & $\pm$\,{\small 2.20}
& 19.46 & $\pm$\,{\small 5.63}
& 6.37 & $\pm$\,{\small 1.01}
& \multicolumn{2}{c}{--}
& 48.38 & $\pm$\,{\small 6.20}
& 23.82 & $\pm$\,{\small 4.87} \\

\textbf{Riksarkivet BK 1}
& 30.20 & $\pm$\,{\small 13.76}
& 21.28 & $\pm$\,{\small 6.84}
& 49.63 & $\pm$\,{\small 9.87}
& 33.40 & $\pm$\,{\small 10.83}
& 38.36 & $\pm$\,{\small 13.85}
& \multicolumn{2}{c}{--}
& 39.29 & $\pm$\,{\small 5.62} \\

\textbf{Riksarkivet GBG 2}
& 55.93 & $\pm$\,{\small 11.09}
& 28.30 & $\pm$\,{\small 11.53}
& 56.26 & $\pm$\,{\small 25.39}
& 38.12 & $\pm$\,{\small 9.48}
& 48.83 & $\pm$\,{\small 1.92}
& 56.19 & $\pm$\,{\small 12.73}
& \multicolumn{2}{c}{--} \\

\midrule


\multicolumn{15}{l}{\textbf{(d) DRAL-HTR (DRA-HTR + Qwen post correction)}} \\
\midrule

\textbf{IAM}
& \multicolumn{2}{c}{--}
& \textbf{3.63} & $\pm$\,{\small 0.12}
& \textbf{15.28} & $\pm$\,{\small 1.19}
& \textbf{4.33} & $\pm$\,{\small 0.18}
& \textbf{14.01} & $\pm$\,{\small 0.52}
& \textbf{17.78} & $\pm$\,{\small 0.48}
& \textbf{17.58} & $\pm$\,{\small 0.13} \\

\textbf{Bentham}
& \textbf{6.61} & $\pm$\,{\small 0.27}
& \multicolumn{2}{c}{--}
& \textbf{19.47} & $\pm$\,{\small 0.72}
& \textbf{4.37} & $\pm$\,{\small 0.10}
& \textbf{15.77} & $\pm$\,{\small 0.28}
& \textbf{16.22} & $\pm$\,{\small 0.19}
& \textbf{20.50} & $\pm$\,{\small 5.50} \\

\textbf{READ 2016}
& \textbf{12.82} & $\pm$\,{\small 0.43}
& \textbf{9.31} & $\pm$\,{\small 0.43}
& \multicolumn{2}{c}{--}
& \textbf{6.59} & $\pm$\,{\small 0.31}
& \textbf{16.27} & $\pm$\,{\small 0.34}
& \textbf{15.59} & $\pm$\,{\small 0.50}
& \textbf{21.30} & $\pm$\,{\small 0.81} \\

\textbf{NorHand 72702}
& \textbf{8.84} & $\pm$\,{\small 0.39}
& \textbf{7.45} & $\pm$\,{\small 2.47}
& \textbf{15.58} & $\pm$\,{\small 0.67}
& \multicolumn{2}{c}{--}
& \textbf{9.34} & $\pm$\,{\small 0.24}
& \textbf{14.60} & $\pm$\,{\small 0.52}
& \textbf{14.72} & $\pm$\,{\small 0.70} \\

\textbf{NorHand 331861}
& \textbf{11.73} & $\pm$\,{\small 0.65}
& \textbf{7.95} & $\pm$\,{\small 0.04}
& \textbf{16.38} & $\pm$\,{\small 1.50}
& \textbf{3.74} & $\pm$\,{\small 0.09}
& \multicolumn{2}{c}{--}
& \textbf{18.33} & $\pm$\,{\small 0.32}
& \textbf{16.09} & $\pm$\,{\small 0.35} \\

\textbf{Riksarkivet BK 1}
& \textbf{15.07} & $\pm$\,{\small 0.26}
& \textbf{7.40} & $\pm$\,{\small 0.17}
& \textbf{17.43} & $\pm$\,{\small 0.37}
& \textbf{6.20} & $\pm$\,{\small 0.11}
& \textbf{19.77} & $\pm$\,{\small 0.65}
& \multicolumn{2}{c}{--}
& \textbf{16.62} & $\pm$\,{\small 0.83} \\

\textbf{Riksarkivet GBG 2}
& \textbf{26.90} & $\pm$\,{\small 9.41}
& \textbf{8.60} & $\pm$\,{\small 3.41}
& \textbf{17.49} & $\pm$\,{\small 0.14}
& \textbf{7.94} & $\pm$\,{\small 2.54}
& \textbf{12.44} & $\pm$\,{\small 0.65}
& \textbf{17.61} & $\pm$\,{\small 6.82}
& \multicolumn{2}{c}{--} \\

\bottomrule
\end{tabular}%
}
\end{table*}

\subsubsection{Impact of DRA-HTR}

To evaluate the effectiveness of our proposed DA, we adapted each HTR model to the six remaining target datasets using our proposed DA pipeline. All experiments were conducted using the same set of hyperparameters, without any dataset-specific tuning. Table~\ref{tab:cross_dataset_comparison}(b) reports the CER obtained after adaptation for each source–target pair.

The results show that our method substantially improves recognition performance on unseen datasets compared with the zero-shot evaluation reported in Table~\ref{tab:cross_dataset_comparison}(a). The average CER across all source–target pairs decreases from 54.39\% without adaptation to 15.39\% with our new pipeline, corresponding to a reduction of 39 percentage points. Furthermore, our method achieves CER values close to 15\% on several target datasets. By leveraging only unlabeled target data, DRA-HTR consistently improves recognition performance across datasets and languages while using a single set of hyperparameters for all adaptation tasks. This enables HTR models to achieve practically usable accuracy on previously unseen datasets by eliminating the requirement of training label creation.

\subsubsection{Qualitative Results}

\begin{figure}[t]
    \centering
    \setlength{\fboxsep}{0pt}
    \setlength{\fboxrule}{0.5pt}

    \begin{minipage}{\linewidth}
        \centering
        \textbf{(a)}\\[0.05cm]
        \fbox{%
            \includegraphics[
                width=\dimexpr\linewidth-2\fboxrule\relax
            ]{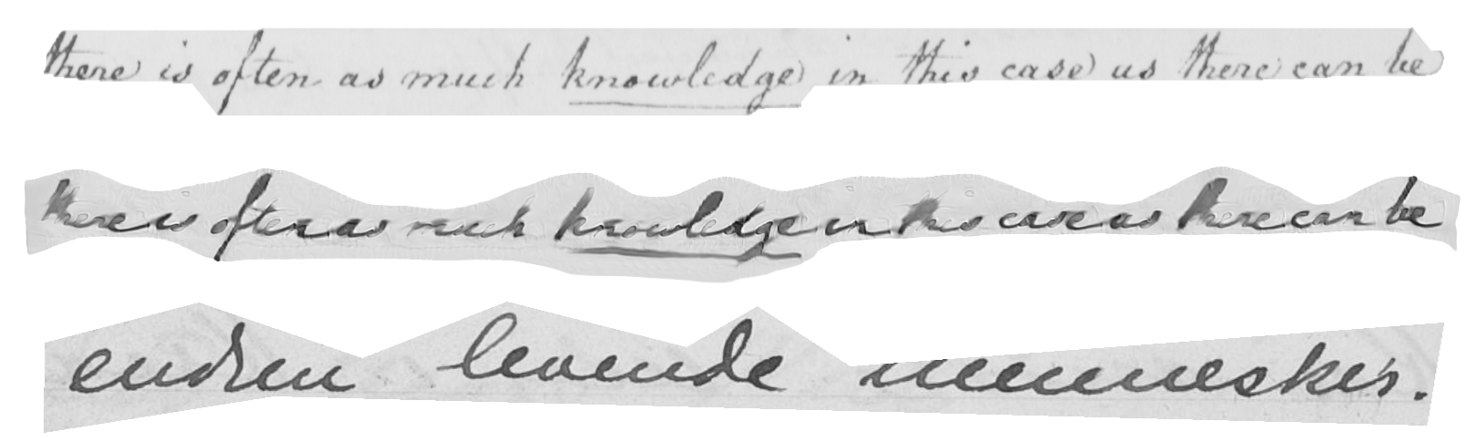}%
        }
    \end{minipage}\\[0.1cm]

    \begin{minipage}{\linewidth}
        \centering
        \textbf{(b)}\\[0.05cm]
        \fbox{%
            \includegraphics[
                width=\dimexpr\linewidth-2\fboxrule\relax
            ]{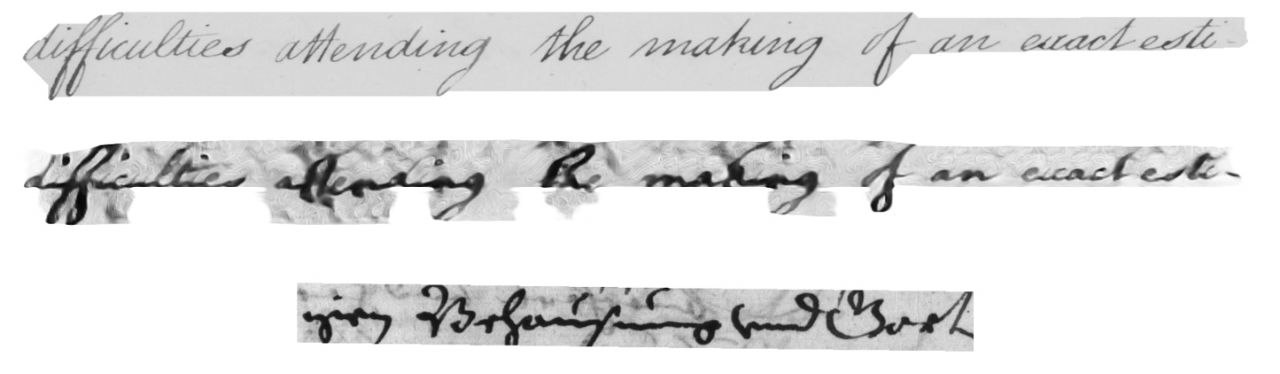}%
        }
    \end{minipage}\\[0.1cm]

    \begin{minipage}{\linewidth}
        \centering
        \textbf{(c)}\\[0.05cm]
        \fbox{%
            \includegraphics[
                width=\dimexpr\linewidth-2\fboxrule\relax
            ]{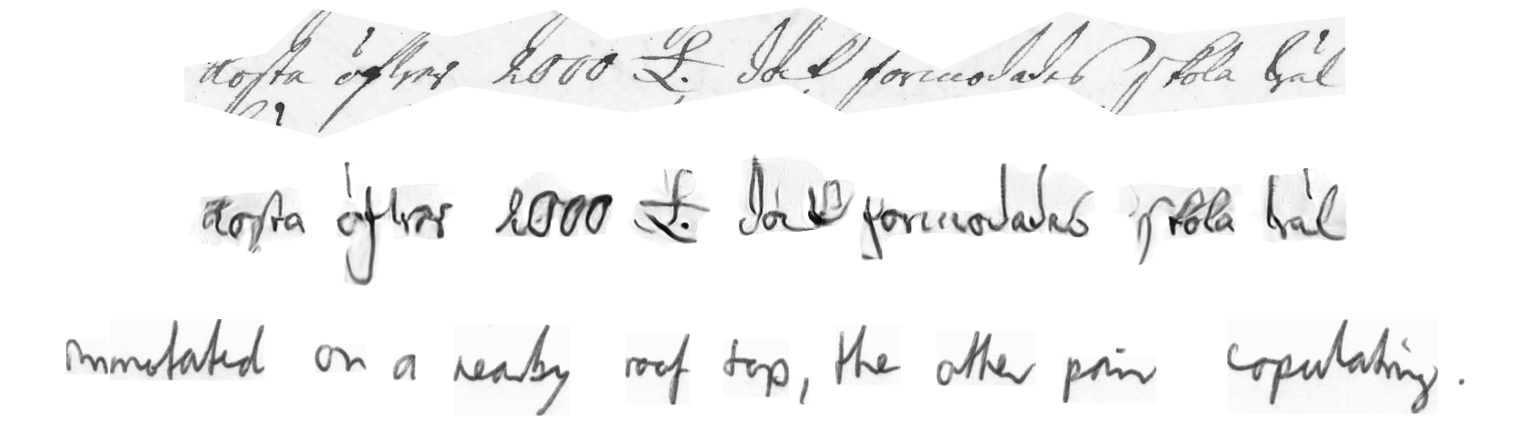}%
        }
    \end{minipage}

    \caption{Three examples of source-to-target style adaptation: (a) Bentham to NorHand 331861, (b) Bentham to READ 2016, and (c) Riksarkivet BK 1 to IAM. In each example, the top row shows a source-domain image, the middle row shows the generated image in the target-domain style, and the bottom row shows a target-domain image.}
    \label{fig:synthetic_image}
\end{figure}

Figure~\ref{fig:synthetic_image} illustrates the source-to-target adaptation process through three representative examples drawn from different datasets. The style adapted samples exhibit a clear shift toward the target-domain appearance: they reproduce target-like background texture, contrast, ink intensity, stroke thickness, and local degradation patterns rather than retaining the visual statistics of the source images.

At the same time, the adaptation preserves the spatial organization of the source text. Character shapes, inter-character spacing, ascenders, descenders, and the overall line geometry remain visually consistent with the source, so that the inherited adaptation remains applicable to the generated image. The adapted lines therefore combine source-domain linguistic content with target-domain visual characteristics, which is precisely the type of labeled target-style data required to train the HTR model.

Overall, the qualitative results indicate that the proposed adaptation framework performs a visually convincing style transfer from source to target while avoiding obvious distortions that would alter the underlying text. 

\subsubsection{Comparison with prior DA}

To evaluate the improvements introduced by our proposed DA pipeline, we compared its performance against the previous method of~\cite{Imbert2026} across all source–target pairs. Table~\ref{tab:cross_dataset_comparison}(c) reports the CER obtained using the previous DA method.

The results show that the previous DA method does improves recognition performance compared with the zero-shot evaluation reported in Table~\ref{tab:cross_dataset_comparison}(a), reducing the average CER by 25.48 percentage points. However, our proposed method further reduces the average CER by 13.52 percentage points, from 28.91\% to 15.39\%. These results demonstrate that the architectural improvements introduced in our pipeline substantially improve recognition performance across different datasets.
Furthermore, our method significantly improves training stability. The average standard deviation across all source–target pairs decreases from 7.23 percentage points with the previous DA method to 1.11 percentage points with our approach, corresponding to a reduction of approximately 85\%. This improvement indicates that our proposed pipeline produces more consistent recognition results across independent training runs, reducing the variability observed with the previous method.

Overall, these results demonstrate that our proposed pipeline improves both recognition accuracy and training stability compared with~\cite{Imbert2026}, making DA more reliable across different source–target configurations.

\subsubsection{Impact of post-HTR correction}

Our DRA-HTR pipeline consists of a three-stage adaptation framework, which can optionally be complemented by Qwen-based post-transcription correction. Table~\ref{tab:cross_dataset_comparison}(d) reports the CER values obtained with additional post-correction. Incorporating Qwen-based post-correction reduces the average CER from \(15.39\%\) to \(13.23\%\), corresponding to an improvement of \(2.16\) percentage points.

\subsubsection{Ablation Study}

In this section, we conduct two ablation studies. The first study investigates the impact of each component of the proposed DA pipeline, while the second examines the choice of the post-HTR correction module. To limit computational costs and GPU resource consumption, both studies are conducted under a single adaptation scenario, using READ 2016 as the source domain and Bentham as the target domain.

\paragraph{Impact of each component}

The contribution of each component is evaluated through the incremental ablation study reported in Table~\ref{tab:abl_cer_comparison}.

\begin{table}[ht]
\centering
\caption{Ablation study of the DA components. E: ConvNeXtV2 encoder (T: Tiny, L: Large) or R for  convolutional component of ResNet-34; P-GAN: PatchGAN discriminator; S\&G: proposed separator and generator; Delay: delayed HTR model training followed by the integration of the task loss; PC: post-HTR correction. Row (a) corresponds to the prior DA method~\cite{Imbert2026}, while row (h) represents the final DRAL-HTR pipeline.}
\label{tab:abl_cer_comparison}
\setlength{\tabcolsep}{3.5pt}
\begin{tabular}{@{}ccccccc@{}}
\toprule
\textbf{} & \textbf{E} & \textbf{P-GAN} & \textbf{S\&G} & \textbf{Delay} & \textbf{PC} & \textbf{CER (\%)} \\
\midrule
(a) & R &            &            &            &            & 22.52 \\
\midrule
(b) & T   &            &            &            &            & 16.57 \\
\midrule
(c) & L   &            &            &            &            & 18.64 \\
\midrule
(d) & T   & \checkmark &            &            &            & 15.53 \\
\midrule
(e) & T   &            & \checkmark &            &            & 20.72 \\
\midrule
(f) & T   & \checkmark & \checkmark &            &            & 16.92 \\
\midrule
(g) & T   & \checkmark & \checkmark & \checkmark &            & 13.53 \\
\midrule
(h) & T   & \checkmark & \checkmark & \checkmark & \checkmark & \textbf{9.31} \\
\bottomrule
\end{tabular}
\end{table}

A first architectural change (b) was replacing the original encoder with ConvNeXtV2-Tiny variant, which substantially improved the visual quality of the generated images and reduced the CER from 22.52\% to 16.57\%. Scaling up to ConvNeXtV2-Large (c) brings no further gain; instead, the CER increases to 18.64\%. A further limitation of the prior method~\mbox{\cite{Imbert2026}} lies in its content--style separation, which occasionally collapses, resulting in generated images that either retain the source-domain appearance or incorporate target-domain textual content along with the target style. Introducing the redesigned separator S and generator G  mitigates these issues and improves training stability across independent runs (e), although at the cost of a 4.15 percentage-point increase in CER.

The redesigned separator and generator are evaluated jointly because their architectures are interdependent. Specifically, the proposed separator produces a compact style vector used for FiLM conditioning, whereas the original generator expects a spatial style map. Evaluating these components independently would therefore require additional adaptation layers, making a direct comparison less representative of the proposed architecture

Replacing the original discriminator with a PatchGAN discriminator (f) further improved the preservation of source text content and reduced the training time. Together, these architectural modifications achieve a CER of 16.92\%. 

Unlike the original approach~\cite{Imbert2026} (a), which does not use recognition-based supervision to guide image generation, we introduce a CTC-based recognition loss to provide an additional supervision signal during DA. However, applying this constraint from the beginning of training may introduce unreliable gradients due to the initially poor quality of the generated images. To address this issue, we adopt a progressive training strategy in which fine-tuning of the HTR model on generated images begins at step \(100{,}000\), while recognition-guided adaptation is activated at step \(130{,}000\). This schedule allows the HTR model to adapt to the synthetic text-line images before its predictions are used to guide the adaptation network. During recognition-guided adaptation, the HTR model remains frozen, while the CTC loss gradients are propagated through the generated images to update the adaptation components. Incorporating this recognition supervision (g) further reduces the CER from \(16.92\%\) to \(13.53\%\), demonstrating the benefit of explicitly enforcing textual consistency during image generation. 

Introducing post-HTR correction (h) further reduces the CER to 9.31\%.

\paragraph{Post-HTR correction}

In this section, we investigate the potential benefits of optional post-transcription correction through a comparative study of five strategies. The results are presented in Table~\ref{tab:lm_comparison}.

\begin{table}[ht]
\centering
\caption{Comparison of post-correction methods for the HTR model adapted from READ 2016 to Bentham. For the Qwen ablation, NP denotes the new prompt, BS beam search, and NL neighboring lines.}
\label{tab:lm_comparison}
\begin{tabularx}{\columnwidth}{@{}c X r@{\,}l@{}}
\toprule
\multicolumn{2}{l}{\textbf{Method}}
& \multicolumn{2}{c}{\textbf{CER (\%)}} \\
\midrule
(a) & Baseline (no post-HTR correction)
    & 13.53 & $\pm$\,{\small 0.40} \\
(b) & CTC beam search
    & 13.37 & $\pm$\,{\small 0.41} \\
(c) & BERT scoring~\cite{shin2019effective}
    & 12.77 & $\pm$\,{\small 0.39} \\
(d) & $n$-gram~\cite{ramirez2022postocr}
    & 10.62 & $\pm$\,{\small 0.22} \\
(e) & Gemini
    & 13.02 & $\pm$\,{\small 0.30} \\
(f) & Qwen~\cite{10.1007/978-3-032-36042-7_14}
    & 11.65 & $\pm$\,{\small 0.52} \\
\midrule
(g) & Qwen with NP
    & 11.37 & $\pm$\,{\small 0.33} \\
(h) & Qwen with NP + NL
    & 10.44 & $\pm$\,{\small 0.45} \\
(i) & Qwen with NP + BS
    & 9.86 & $\pm$\,{\small 0.31} \\
(j) & \textbf{Our}: Qwen with NP + BS + NL 
    & \textbf{9.31} & $\pm$\,{\small 0.29} \\
\bottomrule
\end{tabularx}
\end{table}

CTC beam search (b) explores multiple decoding paths and selects the hypothesis with the highest CTC score. With a beam width of 150, it yields only a marginal improvement. BERT scoring (c) uses the same beam width, retains the top 25 distinct hypotheses, and re-ranks them with multilingual BERT, improving the CER by almost one percentage point over no LM correction (a). Following~\cite{ramirez2022postocr}, the \(n\)-gram approach (d) uses a word-bounded character-level 5-gram model, integrated during decoding through shallow fusion with a beam width of 150, improving the CER by almost three percentage points. For the $n$-gram approach, we train a separate character-level 5-gram language model for each language using KenLM. The training corpora include the Standardized Project Gutenberg Corpus for English, the Nuremberg Letterbooks (2, 3, 4 and 5) and Der Neue Pitaval for German, NorHandv3 for Norwegian, and four publicly available Riksarkivet datasets for Swedish. To prevent data leakage, we exclude books 331861 and 72702 from the Norwegian corpus, and the Riksarkivet BK 1 and GBG 2 collections from the Swedish corpus.

We then evaluate LLM-based post-correction using two language models and two prompt configurations. Configuration~(f) uses Qwen and applying the ``extra-context'' prompt in~\cite{10.1007/978-3-032-36042-7_14} to the greedy CTC output. Gemini~(e) and Qwen~(j) use the prompt provided in Listing~\ref{lst:qwen_post_correction_prompt} and the process in Section~\ref{sec:Language modeling}. While Gemini provides only a marginal improvement, Qwen achieves the largest reduction in CER, improving the result by 4.22 percentage points.

The impact of our proposed improvements to Qwen-based post-HTR correction is evaluated in the last four rows of Table~\ref{tab:lm_comparison}.
The new prompt (g) provides only a slight improvement over (f) but enables the use of additional context. Adding neighboring lines (h) reduces CER by \(0.93\) percentage points, while providing the top 20 beam-search hypotheses (i) reduces it by \(1.51\) points. Combining both strategies (j) achieves the best performance.

When it comes to the computational cost, Qwen requires $2.73~\pm~0.04$\,s per line, compared with $0.25~\pm~0.02$\,s for $n$-gram decoding on an NVIDIA A100 40GB GPU. Despite its higher computational cost, Qwen achieves slightly better recognition accuracy and offers a more general approach that does not require training language-specific models. We therefore adopt Qwen-based post-correction in DRAL-Net to maximize recognition accuracy. However, depending on the application and computational constraints, $n$-gram decoding or no language-based correction can also be considered.

\section{Conclusion and limitation}\label{sec_conclu}

In this paper, we proposed a domain adaptation pipeline specifically designed for HTR by focusing on both visual and language adaptations. The proposed approach achieves strong recognition performance on unseen datasets without requiring annotated target data.

We showed that the proposed DA pipeline substantially improves cross-domain recognition performance compared to previous DA method for HTR~\cite{Imbert2026}, while also providing greater stability across independent training runs. 

Furthermore, we investigated the contribution of post-HTR LLM-based correction in addition to visual DA, resulting in the DRAL-HTR framework. We also validated our three proposed improvements to Qwen-based post-correction: a new prompt, beam search, and the use of neighboring lines. Although post correction improves recognition accuracy, these improvements are mainly observed on English datasets and remain limited for other languages. Moreover, the additional computational cost is relatively high compared to the recognition gains, highlighting a trade-off between accuracy and inference efficiency.

Finally, all experiments were conducted using the same set of hyperparameters across the seven datasets, demonstrating the robustness of the proposed approach. Nevertheless, dataset-specific hyperparameter optimization could further improve recognition performance.

Future work will focus on developing more efficient language adaptation strategies, particularly for non-English datasets, while reducing the computational cost of post-HTR correction.

\bmhead{Acknowledgements}
This work was partially supported by the Wallenberg AI, Autonomous Systems and Software Program (WASP), funded by the Knut and Alice Wallenberg Foundation, with counterpart funding from Luleå University of Technology (LTU).
It was also supported by Kempestiftelserna grant JCSMK23-0109. 
The computations were enabled by the Berzelius resource provided by the Knut and Alice Wallenberg Foundation at the National Supercomputer Centre.

\bibliography{sn-bibliography}

@inproceedings{Retsinas22,
  author       = {George Retsinas and
                  Giorgos Sfikas and
                  Basilis Gatos and
                  Christophoros Nikou},
  title        = {Best Practices for a Handwritten Text Recognition System},
  booktitle    = {Document Analysis Systems - 15th {IAPR} International Workshop, {DAS}
                  2022, Proceedings},
  series       = {Lecture Notes in Computer Science},
  volume       = {13237},
  pages        = {247--259},
  publisher    = {Springer},
  address      = {Cham},
  year         = {2022},
  doi          = {10.1007/978-3-031-06555-2\_17},
  bibsource    = {dblp computer science bibliography, https://dblp.org}
}

@inproceedings{Tarride24,
  author       = {Sol{\`{e}}ne Tarride and
                  Yoann Schneider and
                  Marie Generali{-}Lince and
                  M{\'{e}}lodie Boillet and
                  Bastien Abadie and
                  Christopher Kermorvant},
  title        = {Improving Automatic Text Recognition with Language Models in the {P}y{L}aia
                  Open-Source Library},
  booktitle    = {18th International Conference on Document Analysis and Recognition {ICDAR}, 2024},
  publisher    = {Springer},
  address      = {Cham},
  year         = {2024},
  doi          = {10.1007/978-3-031-70549-6\_23},
  bibsource    = {dblp computer science bibliography, https://dblp.org}
}

@inproceedings{LiL23,
  author       = {Minghao Li and
                  Tengchao Lv and
                  Jingye Chen and
                  Lei Cui and
                  Yijuan Lu and
                  Dinei A. F. Flor{\^{e}}ncio and
                  Cha Zhang and
                  Zhoujun Li and
                  Furu Wei},
  title        = {Tr{OCR}: Transformer-Based Optical Character Recognition with Pre-trained
                  Models},
  booktitle    = {Thirty-Seventh {AAAI} Conference on Artificial Intelligence, {AAAI}
                  2023, Washington, DC, USA
                  7-14, 2023},
  pages        = {13094--13102},
  publisher    = {{AAAI} Press},
  address      = {Washington, DC, USA},
  year         = {2023},
  doi          = {10.1609/AAAI.V37I11.26538},
  bibsource    = {dblp computer science bibliography, https://dblp.org}
}

@article{barrere24,
  TITLE = {{Training transformer architectures on few annotated data: an application to historical handwritten text recognition}},
  AUTHOR = {Barrere, Killian and Soullard, Yann and Lemaitre, Aur{\'e}lie and Co{\"u}asnon, Bertrand},
  URL = {https://hal.science/hal-04447488},
  JOURNAL = {{International Journal on Document Analysis and Recognition}},
  PUBLISHER = {{Springer Verlag}},
  SERIES = {International Journal on Document Analysis and Recognition (IJDAR)},
  VOLUME = {27},
  NUMBER = {4},
  PAGES = {553 - 566},
  YEAR = {2024},
  DOI = {10.1007/s10032-023-00459-2},
  HAL_ID = {hal-04447488},
  HAL_VERSION = {v1},
}

@inproceedings{DRANet,
  author       = {Seunghun Lee and
                  Sunghyun Cho and
                  Sunghoon Im},
  title        = {Disentangling Representation and Adaptation Networks for Unsupervised
                  Cross-Domain Adaptation},
  booktitle    = {{IEEE} Conference on Computer Vision and Pattern Recognition, {CVPR}
                  2021, virtual, June 19-25, 2021},
  pages        = {15252--15261},
  publisher    = {Computer Vision Foundation / {IEEE}},
  address      = {Los Alamitos, CA, USA},
  year         = {2021},
  doi          = {10.1109/CVPR46437.2021.01500},
  bibsource    = {dblp computer science bibliography, https://dblp.org}
}

@incollection{DANN,
  author    = {Ganin, Yaroslav and Ustinova, Evgeniya and Ajakan, Hana and Germain, Pascal and Larochelle, Hugo and Laviolette, Fran{\c{c}}ois and Marchand, Mario and Lempitsky, Victor},
  title     = {Domain-Adversarial Training of Neural Networks},
  booktitle = {Domain Adaptation in Computer Vision Applications},
  pages     = {189--209},
  year      = {2017},
  publisher = {Springer International Publishing},
  address   = {Cham},
  doi       = {10.1007/978-3-319-58347-1_10},
  isbn      = {978-3-319-58347-1}
}

@inproceedings{DeepJDOT,
  author       = {Bharath Bhushan Damodaran and
                  Benjamin Kellenberger and
                  R{\'{e}}mi Flamary and
                  Devis Tuia and
                  Nicolas Courty},
  title        = {Deep{JDOT}: Deep Joint Distribution Optimal Transport for Unsupervised
                  Domain Adaptation},
  booktitle    = {Computer Vision - {ECCV} 2018, Proceedings, Part {IV}},
  series       = {Lecture Notes in Computer Science},
  publisher    = {Springer},
  address      = {Cham},
  year         = {2018},
  doi          = {10.1007/978-3-030-01225-0\_28},
}

@inproceedings{DA_IMU,
  author       = {Florent Imbert and
                  Romain Tavenard and
                  Yann Soullard and
                  {\'{E}}ric Anquetil},
  title        = {Domain Adaptation for Handwriting Trajectory Reconstruction from {IMU}
                  Sensors},
  booktitle    = {Document Analysis and Recognition - {ICDAR} 2024 Workshops - Athens,
                  Greece, August 30-31, 2024, Proceedings, Part {I}},
  series       = {Lecture Notes in Computer Science},
  volume       = {14935},
  pages        = {3--11},
  publisher    = {Springer},
  address      = {Cham},
  year         = {2024},
  doi          = {10.1007/978-3-031-70645-5\_1},
  bibsource    = {dblp computer science bibliography, https://dblp.org}
}

@article{shen2018wassersteindistanceguidedrepresentation,
author = {Shen, Jian and Qu, Yanru and Zhang, Weinan and Yu, Yong},
year = {2018},
month = {04},
pages = {},
title = {Wasserstein Distance Guided Representation Learning for Domain Adaptation},
volume = {32},
journal = {Proceedings of the AAAI Conference on Artificial Intelligence},
doi = {10.1609/aaai.v32i1.11784}
}

@inproceedings{Graves06,
  author       = {Alex Graves and
                  Santiago Fern{\'{a}}ndez and
                  Faustino J. Gomez and
                  J{\"{u}}rgen Schmidhuber},
  title        = {Connectionist temporal classification: labelling unsegmented sequence
                  data with recurrent neural networks},
  booktitle    = {Machine Learning, Proceedings of the Twenty-Third International Conference
                  {(ICML} 2006), USA, 2006},
  series       = {{ACM} International Conference Proceeding Series},
  publisher    = {{ACM}},
  doi          = {10.1145/1143844.1143891},
  bibsource    = {dblp computer science bibliography, https://dblp.org}
}

@article{domain_shift,
author = {Zhou, Kaiyang and Liu, Ziwei and Qiao, Yu and Xiang, Tao and Loy, Chen Change},
title = {Domain Generalization: A Survey},
year = {2023},
issue_date = {April 2023},
publisher = {IEEE Computer Society},
address = {USA},
volume = {45},
number = {4},
issn = {0162-8828},
url = {https://doi.org/10.1109/TPAMI.2022.3195549},
doi = {10.1109/TPAMI.2022.3195549},
journal = {IEEE Trans. Pattern Anal. Mach. Intell.},
month = apr,
pages = {4396–4415},
numpages = {20}
}

@inproceedings{SELayer,
  author       = {Jie Hu and
                  Li Shen and
                  Gang Sun},
  title        = {Squeeze-and-Excitation Networks},
  booktitle    = {2018 {IEEE} Conference on Computer Vision and Pattern Recognition,
                  {CVPR} 2018, Salt Lake City, UT, USA, June 18-22, 2018},
  pages        = {7132--7141},
  publisher    = {Computer Vision Foundation / {IEEE} Computer Society},
  address      = {Los Alamitos, CA, USA},
  year         = {2018},
  doi          = {10.1109/CVPR.2018.00745},
  bibsource    = {dblp computer science bibliography, https://dblp.org}
}

@INPROCEEDINGS{Corbille25,
  author={Corbillé, Simon and Barney Smith, Elisa H},
  booktitle={2025 International Joint Conference on Neural Networks (IJCNN)}, 
  title={Applying Center Loss to Neural Networks for Sequence Prediction: A Study for Handwriting Recognition}, 
  year={2025},
  volume={},
  number={},
  pages={1-7},

  doi={10.1109/IJCNN64981.2025.11227321}}

@article{Imbert2026,
  author  = {Imbert, Florent and Corbill{\'e}, Simon and Han, Hui and Barney Smith, Elisa H.},
  title   = {Domain Adaptation Based Pipeline for Character Classification and Handwritten Text Recognition},
  journal = {International Journal on Document Analysis and Recognition (IJDAR)},
  volume  = {29},
  year    = {2026},
  doi     = {10.1007/s10032-026-00575-9},
  url     = {https://doi.org/10.1007/s10032-026-00575-9},
  issn    = {1433-2825}
}

@InProceedings{ConvNeXtV2,
    author    = {Woo, Sanghyun and Debnath, Shoubhik and Hu, Ronghang and Chen, Xinlei and Liu, Zhuang and Kweon, In So and Xie, Saining},
    title     = {{ConvNeXt} V2: Co-Designing and Scaling {ConvNets} With Masked Autoencoders},
    booktitle = {Proceedings of the IEEE/CVF Conference on Computer Vision and Pattern Recognition (CVPR)},
    month     = {June},
    year      = {2023},
}

@article{COQUENET2026113373,
title = {Meta-DAN: Towards an efficient prediction strategy for page-level handwritten text recognition},
journal = {Pattern Recognition},
volume = {177},
pages = {113373},
year = {2026},
issn = {0031-3203},
doi = {https://doi.org/10.1016/j.patcog.2026.113373},
url = {https://www.sciencedirect.com/science/article/pii/S0031320326003389},
author = {Denis Coquenet}
}

@InProceedings{10.1007/978-3-032-36042-7_14,
author="Koynov, Radoslav
and Doan, Triet Ho Anh
and Wieder, Philipp",
title="Vision Language Models as {OCR} Correctors for Historical Texts",
booktitle="Document Analysis and Recognition -- ICDAR 2026",
year="2027",
publisher="Springer Nature Switzerland",
address="Cham",
pages="229--246",
isbn="978-3-032-36042-7",
doi = "https://doi.org/10.1007/978-3-032-36042-7_14"
}

@inproceedings{shin2019effective,
  title={Effective sentence scoring method using {BERT} for speech recognition},
  author={Shin, Joonbo and Lee, Yoonhyung and Jung, Kyomin},
  booktitle={Asian Conference on Machine Learning},
  pages={1081--1093},
  year={2019},
  organization={PMLR},
}

@inproceedings{salazar2020masked,
    title = "Masked Language Model Scoring",
    author = "Salazar, Julian  and
      Liang, Davis  and
      Nguyen, Toan Q.  and
      Kirchhoff, Katrin",
    booktitle = "Proceedings of the 58th Annual Meeting of the {A}ssociation for {C}omputational {L}inguistics",
    month = jul,
    year = "2020",
    address = "Online",
    publisher = "Association for Computational Linguistics",
    doi = "10.18653/v1/2020.acl-main.240",
    pages = "2699--2712"
}

@InProceedings{meyer2026n,
author="Meyer, Florent
and Guichard, Laurent
and Soullard, Yann
and Coquenet, Denis
and Gravier, Guillaume
and Co{\"u}asnon, Bertrand",
editor="Fink, Gernot A.
and Forn{\'e}s, Alicia
and Kise, Koichi
and Lopresti, Daniel",
title="n-Gram Injection into Transformers for Dynamic Language Model Adaptation in Handwritten Text Recognition",
booktitle="Document Analysis and Recognition -- ICDAR 2026",
year="2027",
publisher="Springer Nature Switzerland",
address="Cham",
pages="572--589",
isbn="978-3-032-36033-5",
doi= {10.1007/978-3-032-36033-5_34}
}

@InProceedings{Gu_2025_CVPR,
    author    = {Gu, Wenhao and Gu, Li and Suen, Chingyee Yee and Wang, Yang},
    title     = {MetaWriter: Personalized Handwritten Text Recognition Using Meta-Learned Prompt Tuning},
    booktitle = {Proceedings of the IEEE/CVF Conference on Computer Vision and Pattern Recognition (CVPR)},
    month     = {June},
    year      = {2025},
}

@article{Chinthaginjala2024,
  author  = {Chinthaginjala, Ravikumar and Dhanamjayulu, C. 
             and Kim, Tai-hoon and Ahmed, Suhaib 
             and Kim, Si-Yeong and Kumar, A. S. 
             and Annepu, Visalakshi and Ahmad, Shafiq},
  title   = {Enhancing Handwritten Text Recognition Accuracy 
             with Gated Mechanisms},
  journal = {Scientific Reports},
  volume  = {14},
  number  = {1},
  pages   = {16800},
  year    = {2024},
  doi     = {10.1038/s41598-024-67738-8}
}

@ARTICLE{DAN,
  author={Coquenet, Denis and Chatelain, Clément and Paquet, Thierry},
  journal={IEEE Transactions on Pattern Analysis and Machine Intelligence}, 
  title={DAN: A Segmentation-Free Document Attention Network for Handwritten Document Recognition}, 
  year={2023},
  volume={45},
  number={7},
  pages={8227-8243},
  doi={10.1109/TPAMI.2023.3235826}}

@inproceedings{Perez2018FiLM,
  title={Film: Visual reasoning with a general conditioning layer},
  author={Perez, Ethan and Strub, Florian and De Vries, Harm and Dumoulin, Vincent and Courville, Aaron},
  booktitle={Proceedings of the AAAI conference on artificial intelligence},
  volume={32},
  number={1},
  year={2018},
  doi={https://doi.org/10.1609/aaai.v32i1.11671}
}

@inproceedings{pathirage2026generalization,
  author    = {Pathirage, Gayan H. and Unter, Stephan M. and
               Corbillé, Simon and Barney Smith, Elisa H.},
  title     = {Generalization of Text Line Segmentation for
               {HTR} in Historical Documents},
  booktitle = {Document Analysis and Recognition -- ICDAR 2026
               Workshops,
               Proceedings},
  series    = {Lecture Notes in Computer Science},
  volume    = {16983},
  publisher = {Springer},
  address   = {Cham}
}

@InProceedings{Ren2021QSpace,
    author="Ren, Mengwei
    and Kim, Heejong
    and Dey, Neel
    and Gerig, Guido",
    title="Q-space Conditioned Translation Networks for Directional Synthesis of Diffusion Weighted Images from Multi-modal Structural {MRI}",
    booktitle="Medical Image Computing and Computer Assisted Intervention -- MICCAI 2021",
    year="2021",
    publisher="Springer International Publishing",
    address="Cham",
    pages="530--540",
    doi={10.1007/978-3-030-87234-2_50}
}

@ARTICLE{Ren2021Segmentation,
  author={Ren, Mengwei and Dey, Neel and Fishbaugh, James and Gerig, Guido},
  journal={IEEE Transactions on Medical Imaging}, 
  title={Segmentation-Renormalized Deep Feature Modulation for Unpaired Image Harmonization}, 
  year={2021},
  volume={40},
  number={6},
  pages={1519-1530},
  doi={10.1109/TMI.2021.3059726}
}

@misc{NorHandv3,
  author       = {Beyer, Yngvil and
                  Solberg, Per Erik},
  title        = {{NorHand v3 / Dataset for Handwritten Text
                   Recognition in Norwegian}},
  month        = dec,
  year         = 2023,
  publisher    = {Zenodo},
  doi          = {10.5281/zenodo.10255840},
}

@article{Marti2002IAM,
    author    = {Marti, U.-V. and Bunke, H.},
    title     = {The {IAM}-database: An {English} Sentence Database for Offline Handwriting Recognition},
    journal   = {International Journal on Document Analysis and Recognition},
    volume    = {5},
    number    = {1},
    pages     = {39--46},
    year      = {2002},
    doi       = {10.1007/s100320200071},
    url       = {https://doi.org/10.1007/s100320200071}
}

@INPROCEEDINGS{Sanchez2014Bentham,
  author={Sánchez, Joan Andreu and Romero, Verónica and Toselli, Alejandro H. and Vidal, Enrique},
  booktitle={2014 14th International Conference on Frontiers in Handwriting Recognition}, 
  title={{ICFHR2014} Competition on Handwritten Text Recognition on {Transcriptorium} Datasets {(HTRtS)}}, 
  year={2014},
  volume={},
  number={},
  pages={785-790},
  doi={10.1109/ICFHR.2014.137}}

@INPROCEEDINGS{Sanchez2016READ,
  author={Sánchez, Joan Andreu and Romero, Verónica and Toselli, Alejandro H. and Vidal, Enrique},
  booktitle={2016 15th International Conference on Frontiers in Handwriting Recognition {(ICFHR)}}, 
  title={{ICFHR2016} Competition on Handwritten Text Recognition on the {READ} Dataset}, 
  year={2016},
  volume={},
  number={},
  doi={10.1109/ICFHR.2016.0120}}

@inproceedings{boros-etal-2024-post,
    title = "Post-Correction of Historical Text Transcripts with Large Language Models: An Exploratory Study",
    author = "Boros, Emanuela  and
      Ehrmann, Maud  and
      Romanello, Matteo  and
      Najem-Meyer, Sven  and
      Kaplan, Fr{\'e}d{\'e}ric",
    booktitle = "Proceedings of the 8th Joint {SIGHUM} Workshop on Computational Linguistics for Cultural Heritage, Social Sciences, Humanities and Literature (2024)",
    publisher = "Association for Computational Linguistics",
    doi = "10.18653/v1/2024.latechclfl-1.14",
}

@misc{alfaro2024align,
      title={Align, Minimize and Diversify: A Source-Free Unsupervised Domain Adaptation Method for Handwritten Text Recognition}, 
      author={María Alfaro-Contreras and Jorge Calvo-Zaragoza},
      year={2024},
      eprint={2404.18260},
      archivePrefix={arXiv},
}

@article{ramirez2022postocr,
  title={Post-OCR Document Correction with Large Ensembles of Character Sequence-to-Sequence Models},
  author={Ramirez-Orta, Juan Antonio and Xamena, Eduardo and Maguitman, Ana Gabriela and Milios, Evangelos E. and Soto, Axel J.},
  journal={Proceedings of the AAAI Conference on Artificial Intelligence},
  volume={36},
  number={10},
  pages={11192--11199},
  year={2022},
  doi={10.1609/aaai.v36i10.21369}
}

\end{document}